\documentclass[letterpaper]{article} % DO NOT CHANGE THIS
\usepackage{aaai25}  % DO NOT CHANGE THIS
\usepackage{times}  % DO NOT CHANGE THIS
\usepackage{helvet}  % DO NOT CHANGE THIS
\usepackage{courier}  % DO NOT CHANGE THIS
\usepackage[hyphens]{url}  % DO NOT CHANGE THIS
\usepackage{graphicx} % DO NOT CHANGE THIS
\usepackage{natbib}  % DO NOT CHANGE THIS AND DO NOT ADD ANY OPTIONS TO IT
\usepackage{caption} % DO NOT CHANGE THIS AND DO NOT ADD ANY OPTIONS TO IT
\usepackage{algorithm}
\usepackage{algorithmic}
\usepackage{newfloat}
\usepackage{listings}
\DeclareCaptionStyle{ruled}{labelfont=normalfont,labelsep=colon,strut=off} % DO NOT CHANGE THIS
\floatstyle{ruled}
\newfloat{listing}{tb}{lst}{}
\floatname{listing}{Listing}
\usepackage{amssymb}
\usepackage{booktabs}
\usepackage{multirow}
\usepackage{arydshln}
\usepackage{overpic}
\usepackage{array}
\usepackage{dsfont}
\usepackage[caption=false,font=normalsize,labelfont=sf,textfont=sf]{subfig}
\usepackage{colortbl} %彩色表格需要加载的宏包
\usepackage{xcolor}
\usepackage{subcaption, textpos}
\usepackage{amsmath,amsfonts}
\usepackage{url}
\usepackage{textcomp}
\usepackage{verbatim}
\newlength\savewidth\newcommand\shline{\noalign{\global\savewidth\arrayrulewidth
  \global\arrayrulewidth 1pt}\hline\noalign{\global\arrayrulewidth\savewidth}}

\renewcommand{\paragraph}[1]{\vspace{1.25mm}\noindent\textbf{#1}}

\newcommand{\app}{\raise.17ex\hbox{$\scriptstyle\sim$}}

\definecolor{deemph}{gray}{0.6}

\definecolor{baselinecolor}{gray}{.9}

\definecolor{color4}{rgb}{0.94,0.94,1}
\title{Intra and Inter Parser-Prompted Transformers for Effective Image Restoration}
\author{
    Cong Wang\textsuperscript{\rm 1}\textsuperscript{\rm 2}\textsuperscript{\rm 3}
    Jinshan Pan\textsuperscript{\rm 4},
    Liyan Wang\textsuperscript{\rm 5},
    Wei Wang\textsuperscript{\rm 1}\thanks{Wei Wang is the corresponding author.}
}
\affiliations{
    \textsuperscript{\rm 1}Shenzhen Campus of Sun Yat-sen University, China\\
    \textsuperscript{\rm 2}Centre for Advances in Reliability and Safety, Hong Kong\\
    \textsuperscript{\rm 3}The Hong Kong Polytechnic University, Hong Kong\\
    \textsuperscript{\rm 4}Nanjing University of Science and Technology, China\\
    \textsuperscript{\rm 5}Dalian University of Technology, China\\
    supercong94@gmail.com, sdluran@gmail.com, wangwei29@mail.sysu.edu.cn
}
\usepackage{bibentry}
\begin{document}
\maketitle
\begin{abstract}
We propose Intra and Inter Parser-Prompted Transformers (PPTformer) that explore useful features from visual foundation models for image restoration. Specifically, PPTformer contains two parts: an Image Restoration Network (IRNet) for restoring images from degraded observations and a Parser-Prompted Feature Generation Network (PPFGNet) for providing IRNet with reliable parser information to boost restoration. To enhance the integration of the parser within IRNet, we propose Intra Parser-Prompted Attention (IntraPPA) and Inter Parser-Prompted Attention (InterPPA) to implicitly and explicitly learn useful parser features to facilitate restoration. The IntraPPA re-considers cross attention between parser and restoration features, enabling implicit perception of the parser from a long-range and intra-layer perspective. Conversely, the InterPPA initially fuses restoration features with those of the parser, followed by formulating these fused features within an attention mechanism to explicitly perceive parser information. Further, we propose a parser-prompted feed-forward network to guide restoration within pixel-wise gating modulation. Experimental results show that PPTformer achieves state-of-the-art performance on image deraining, defocus deblurring, desnowing, and low-light enhancement. 
\end{abstract}
\begin{links}
\link{Code}{https://github.com/supersupercong/pptformer}
% \link{Datasets}{https://aaai.org/example/datasets}
% \link{Extended version}{https://aaai.org/example/extended-version}
\end{links}
% The source codes will be made available at https://github.com/supersupercong/pptformer.

% \vspace{-3mm}
\section{Introduction}
% \vspace{-1mm}
Image restoration aims to reconstruct high-quality images from their degraded counterparts. 
This task is inherently challenging because it relies solely on the degraded images themselves while the clear images and degradation factors are unknown. 
To effectively solve this problem, statistical observations are employed to transform it into a `well-posed' one, as discussed in various studies~\cite{pan2016blind}. 
While traditional methods have achieved some success in restoration, they are hampered by complex optimization algorithms that struggle with issues of non-convexity and non-smoothness. 
%
%Further, the underlying assumptions made by these methods do not always hold on, leading to potential failures in the restoration process.

\begin{figure}[!t]
% \vspace{-8mm}
% \footnotesize
\centering
\begin{center}
\begin{tabular}{c}
\includegraphics[width=1\linewidth]{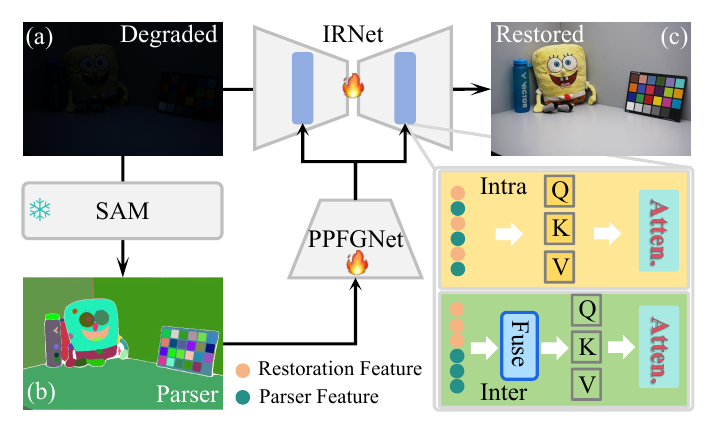} 
\end{tabular}
\vspace{-2mm}
\caption{Illustration of our main idea.
Our method is motivated by an interesting observation that SAM~\cite{kirillov2023segany} can parse degraded images into useful hierarchical structures (b) although severely degraded inputs (a).
While extreme degradation may not provide valuable information for restoration, the parser that benefits from the powerful ability of SAM can still describe reliable structures well to facilitate restoration.
To better integrate the parser into the restoration process, we develop Intra Parser-Prompted Attention and Inter Parser-Prompted Attention to implicitly and explicitly learn valuable parser content to boost image recovery.
}
\label{fig: Illustration of our main idea.}
\end{center}
% \vspace{-5mm}
\end{figure}

The emergence of convolutional neural networks~\cite{he2016deep} and Transformers~\cite{vision_transformer} has revolutionized image restoration tasks. 
These advanced models have excelled by implicitly learning from large-scale data. 
This learning-based strategy has taken precedence in contemporary image restoration, outperforming previous methods with notable success, as evidenced by a series of studies~\cite{ren2019progressive,liu2018desnownet,wang_aaai22,dcsfn-wang-mm20,jdnet-wang-mm20,wang2021uformer,guo2022dehamer,peng2024towards,peng2024lightweight,wang2024promptrestorer,wang2024selfpromer,wang2024uhdformer,wang2024msgnn,wang2024progressive,wang2024perceplie,xu2024beyond,xu2024end}. 
However, we note that most of the existing state-of-the-art approaches learn the `self' degraded knowledge without considering additional helpful knowledge from other domains into network designs, limiting model capacity to some extent.

To address this problem, some conditional modulation-driven networks are suggested~\cite{condition_Wang_2018_CVPR}.
These approaches usually contain a restoration branch for image reconstruction and a conditional branch for providing the restoration branch with useful conditional information.
Among them, degraded images are usually severed as conditions to provide pixel-wise knowledge~\cite{condition_Wang_2018_CVPR}.
We notice that the degraded images usually contain unreliable pixels due to the degraded disruption.
For example, in low-light conditions, extensive pixels approximate zero (as illustrated in Fig.~\ref{fig: Illustration of our main idea.}(a)), which may not provide reliable content for the restoration branch, thus limiting recovery performance. 

Recent large visual foundation models, e.g., SAM~\cite{kirillov2023segany}, have shown strong ability in visual understanding~\cite{lu2023can}.
As one of the powerful visual foundation models, the SAM model can effectively parse degraded images into hierarchical structure content although severe degradation (see Fig.~\ref{fig: Illustration of our main idea.}(b)).
Compared with degraded images (Fig.~\ref{fig: Illustration of our main idea.}(a)) which almost cannot provide useful information, the parsed contents (Fig.~\ref{fig: Illustration of our main idea.}(b)) contain semantic information with salient structures.
%, which may be helpful for restoration.
%
Therefore, \textit{a natural question is whether the parsed contents help image restoration. If so, how do we explore the parsed contents to better help image restoration?}

In this paper, we propose an Intra and Inter Parser-Prompted Transformer (PPTformer) to answer the above questions.
Our PPTformer adequately considers the parsing content generated by SAM~\cite{kirillov2023segany} into the restoration process within both long-range pixel dependency and pixel-wise modulation perspectives.
Specifically, our PPTformer contains two parts: one is the Image Restoration Network (IRNet) for restoring images from the degraded observations while another is the Parser-Prompted Feature Generation Network (PPFGNet) for providing IRNet with reliable parsing features to boost restoration performance.
To better integrate parsing content into IRNet, we propose the Intra Parser-Prompted Attention (IntraPPA) and the Inter Parser-Prompted Attention (InterPPA), which implicitly and explicitly learn useful parser features to guide restoration, respectively.
The IntraPPA re-considers cross attention between parser features and restoration ones to implicitly perceive the parser content within the long-range intra-layer perspective, while the InterPPA first fuses the restoration features with parser ones and then conducts attention computation to explicitly perceive the parser features.
Further, we propose a Parser-Prompted Feed-forward Network (PPFN) to guide restoration from a pixel-wise gating modulation perspective.
Moreover, we propose a bidirectional parser-prompted fusion scheme to better fuse the parser features and restoration ones, which would be adopted in both InterPPA and PPFN.
Fig.~\ref{fig: Illustration of our main idea.} illustrates the main idea of our PPTformer.

The main contributions are summarized below:
% \begin{itemize}
\begin{itemize}
     \item We propose an intra and inter parser-prompted Transformer, which integrates the visual foundation models into the restoration process.
    \item We propose an intra parser-prompted attention and an inter parser-prompted attention to implicitly and explicitly explore useful parser features to facilitate restoration.
    \item We suggest a bidirectional parser-prompted fusion scheme to effectively fuse parser and restoration features.

    \item We show that the proposed method achieves favorable results on image restoration tasks including image deraining, single-image defocus deblurring, image desnowing, and low-light image enhancement.
\end{itemize}
% \end{itemize}

\begin{figure*}[!t]
% \scriptsize
\centering
\begin{center}
\begin{tabular}{c}
\includegraphics[width=0.99\linewidth]{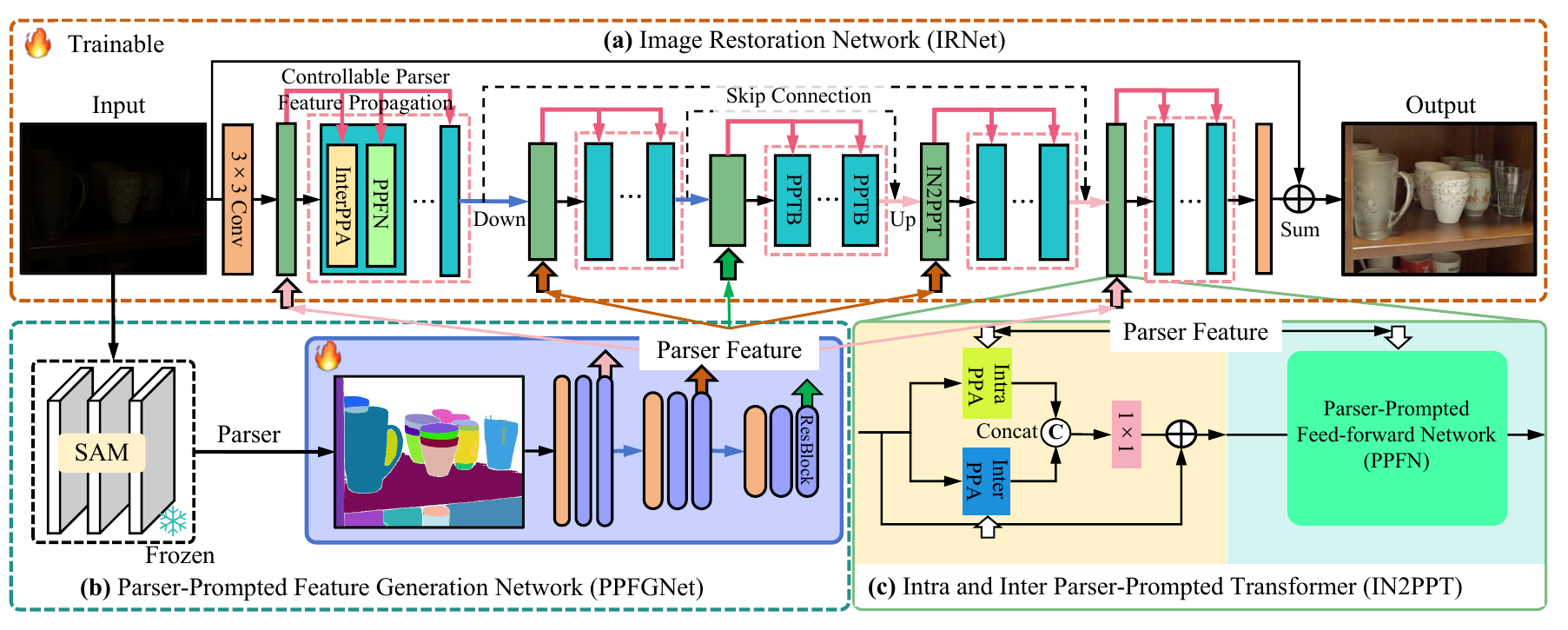} 
\end{tabular}
\vspace{-2mm}
\caption{Overall framework of PPTformer.
Our PPTformer consists of two parts: \textbf{(a)} Image Restoration Network (IRNet); \textbf{(b)} Parser-Prompted Feature Generation Network (PPFGNet).
The IRNet is used to restore images, while PPFGNet is used to generate parser features to provide IRNet with useful information to facilitate restoration.
To better utilize the parser features to guide IRNet, we propose the Intra and Inter Parser-Prompted Attention, which implicitly and explicitly explore the useful parser features in the restoration process.
Further, we propose the Parser-Prompted Feed-forward Network to integrate parser features into the feed-forward encoding process, which allows parser features to effectively guide the restoration within the pixel-wise gating modulation perspective.
Moreover, we introduce the Controllable Parser Feature Propagation scheme to control parser feature propagation in both attention and networks to allow useful information to be passed for better guide image restoration.
}
\label{fig: Overall framework of our MPTIR}
\end{center}
% \vspace{-5mm}
\end{figure*}

\section{Proposed Approach}

Our goal aims to exploit SAM~\cite{kirillov2023segany} to parse degraded images into hierarchical structures to prompt the restoration process with more useful information to facilitate restoration. 
To that end, we propose the PPTformer, an intra and inter Parser-Prompted Transformer. 
To better integrate parser information into restoration, we propose an intra parser-prompted attention, an inter parser-prompted attention, and a parser-prompted feed-forward network.
We also suggest a bidirectional parser-prompted fusion scheme to fuse parser features with restoration ones.

\subsection{Overall pipeline}\label{sec:Overall Pipeline}
\begin{figure*}[!t]\scriptsize
\centering
\begin{center}
\begin{tabular}{c}
\includegraphics[width=1\linewidth]{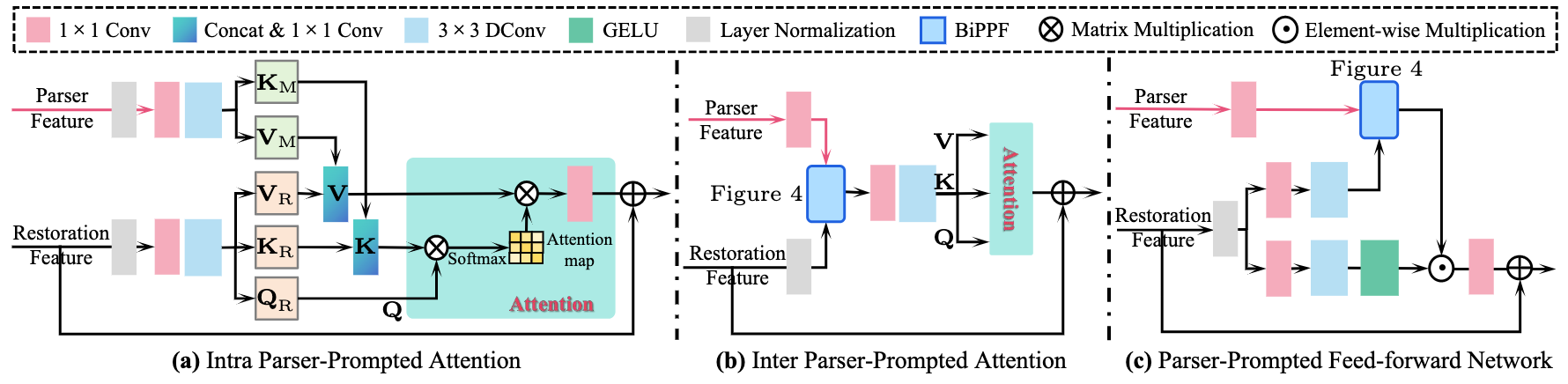} 
\end{tabular}
\vspace{-2mm}
\caption{(a) Intra Parser-Prompted Attention (IntraPPA), (b) Inter Parser-Prompted Attention (InterPPA), and (c) Parser-Prompted Feed-forward Network (PPFN).
Our IntraPPA exploits the cross-attention between parser features and restoration features to implicitly explore useful parser features.
The InterPPA explicitly explores the aggregation, which first utilizes BiPPF (see Fig.~\ref{fig: Bidirectional Mask-Prompted Fusion}) to fuse parser features with restoration ones and then conducts attention to explicitly learn beneficial parser features.
%
% PPN utilizes the gating mechanism by the guidance of parser features within the bridge of BiMPF.
The PPFN integrates parser features into one of the parallel paths by BiPPF, which allows parser features to effectively guide the feed-forward restoration process within a pixel-wise gating modulation mechanism.
}
\label{fig: IntraMPA_InterMPA_MPN}
\end{center}
% \vspace{-5mm}
\end{figure*}

Fig.~\ref{fig: Overall framework of our MPTIR} shows the overview of our PPTformer. It contains two parts: \textbf{(a)} Image Restoration Network (IRNet) and \textbf{(b)} Parser-Prompted Feature Generation Network (PPFGNet).
The IRNet is used to restore images from given degraded observations while the PPFGNet is used to parse input degraded images into hierarchical structures to prompt the IRNet with useful information to facilitate restoration.
To better fuse parser features into IRNet, we propose Intra Parser-Prompted Attention (see Fig.~\ref{fig: IntraMPA_InterMPA_MPN}(a)), Inter Parser-Prompted Attention (see Fig.~\ref{fig: IntraMPA_InterMPA_MPN}(b)) to implicitly and explicitly learn useful parser features from the long-range pixel dependency perspective, respectively.
Furthermore, we propose a Parser-Prompted Feed-forward Network (see Fig.~\ref{fig: IntraMPA_InterMPA_MPN}(c)) to integrate parser features into the restoration process within the pixel-wise gating modulation perspective.
Moreover, we propose a Bidirectional Parser-Prompted Fusion scheme to effectively fuse parser features and restoration ones.
{\flushleft \bf Image Restoration Network.}
Given a degraded input image $\mathbf{I}$~$\in$~$\mathbb{R}^{H\times W \times 3}$, we first apply a $3 \times 3$ convolution as the feature extraction to obtain low-level embeddings $\mathbf{X}_0$~$\in$~$\mathbb{R}^{H\times W \times C}$; where $H\times W$ denotes the spatial dimension and $C$ is the number of channels. 
Next, the shallow features $\mathbf{X}_{0}$ gradually are hierarchically encoded into deep features $\mathbf{X}_l$~$\in$~$\mathbb{R}^{\frac{H}{l}\times \frac{W}{l} \times lC}$.
After encoding the degraded input into low-resolution latent features $\mathbf{X}_{3}$~$\in$~$\mathbb{R}^{\frac{H}{3}\times\frac{W}{3}\times 3C}$, the decoder progressively recovers the high-resolution representations. 
Finally, a reconstruction layer which contains a $3 \times 3$ convolution is applied to decoded features to generate residual image $\mathbf{S}$~$\in$~$\mathbb{R}^{H\times W \times 3}$ to which degraded image is added to obtain the restored output image: $\mathbf{\hat{H}} = \mathbf{I} + \mathbf{S}$. 

Both encoder and decoder at $l$- level consists of one Inter and Intra Parser-Prompted Transformer (IN2PPT), as shown in Fig.~\ref{fig: Overall framework of our MPTIR}(c), to perceive parser features, and multiple Parser-Prompted Transformers Blocks (PPTB) which consists of a Parser-Prompted Attention (PPA) and a Parser-Prompted Feed-forward Network (PPFN).
% %
To help better recovery, the encoder features are concatenated with decoder features via skip connections~\cite{ronneberger2015unet} by $1$$\times$$1$ convolutions. 

{\flushleft \bf Parser-Prompted Feature Generation Network.}
The PPFGNet aims to parse degraded images into hierarchical structures and then generates multi-scale parser features to provide guidance for IRNet.
Specifically, we first utilize SAM~\cite{kirillov2023segany} to parse input images into hierarchical structures, which are then input to an autoencoder-like network to generate multi-scale parser features.
The parser features would be utilized to prompt IRNet with useful information to facilitate better restoration.

%
% \vspace{-1mm}
\subsection{Intra and Inter Parser-Prompted Transformer}\label{sec: Inter and Intra Mask-Prompted Transformer}
% \vspace{-1mm}
%
\begin{figure}[!t]
\vspace{-1.5mm}
\footnotesize
\centering
\begin{center}
\begin{tabular}{c}
\includegraphics[width=1\linewidth]{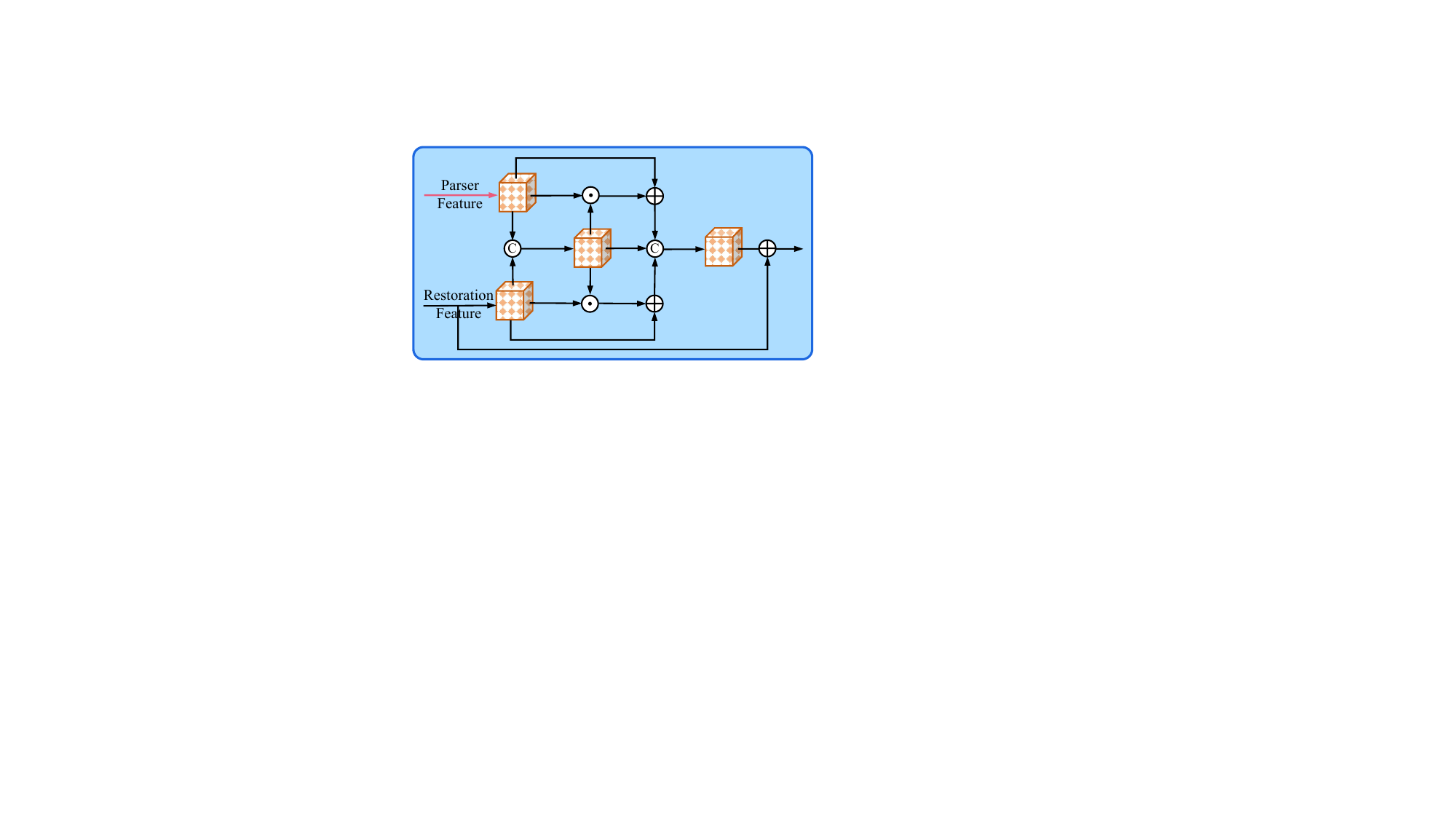} 
\end{tabular}
\vspace{-2mm}
\caption{Bidirectional Parser-Prompted Fusion (BiPPF).
Our BiPPF method effectively integrates a bidirectional flow scheme for feature fusion. It transforms parser features into restoration features and vice versa, allowing interactive integration of parser features into the restoration process.
Notably, all convolutions in BiPPF are 1$\times$1 for efficient design.
}
\label{fig: Bidirectional Mask-Prompted Fusion}
\end{center}
% \vspace{-5mm}
\end{figure}
To effectively perceive and exploit parser features, we propose the Intra and Inter Parser-Prompted Transformer (IN2PPT), which contains an Intra Parser-Prompted Attention (IntraPPA), an Inter Parser-Prompted Attention (InterPPA), and a Parser-Prompted Feed-forward Network (PPFN).
The IntraPPA, shown in Fig.~\ref{fig: IntraMPA_InterMPA_MPN}(a), implicitly learns parser features by building cross-attention between parser features and restoration ones, while the InterPPA, shown in Fig.~\ref{fig: IntraMPA_InterMPA_MPN}(b), explicitly explores useful parser features by first fusing parser features via Bidirectional Parser-Prompted Fusion (see Fig.~\ref{fig: Bidirectional Mask-Prompted Fusion}) and restoration ones before conducting attention computation, both to prompt restoration from the long-range pixel dependency perspective.
The PPFN, shown in Fig.~\ref{fig: IntraMPA_InterMPA_MPN}(c), integrates parser features into restoration by pixel-wise gating modulation.
The IntraPPA and InterPPA adopt the parallel structures and are further concatenated and fused by one 1$\times$1 convolution.
The fused features are then sent to PPFN.

{\flushleft \bf Intra Parser-Prompted Attention.}
The IntraPPA, shown in Fig.~\ref{fig: IntraMPA_InterMPA_MPN}(a), fully considers the implicit global fusion between parser features and restoration ones.
IntraPPA utilizes cross-attention to interactively learn more useful parser features for restoration, allowing restoration features to implicitly explore more reliable parser information to effectively guide the restoration process.
Specifically, we first split parser features $\mathbf{M}$ into two tensors: Key ($\textbf{K}_{\textbf{M}}$) and Value ($\textbf{V}_{\textbf{M}}$), by a 1$\times$1 point-wise convolution ($W_p$) and 3$\times$3 depth-wise convolution ($W_d$).
With similar manner, restoration features $\mathbf{X}$ are split into three tensors: Query ($\textbf{Q}_{\textbf{R}}$), Key ($\textbf{K}_{\textbf{R}}$), and Value ($\textbf{V}_{\textbf{R}}$).
Then, we respectively fuse the Keys and Values of restoration features and parser features by concatenation and 1$\times$1 convolution to form a new Key and Value.
Lastly, the Query, Key, and Value vectors are performed attention.
The process of IntraPPA can be formulated by:
%
% \vspace{-2mm}
\begin{equation}
% \vspace{-2mm}
\begin{split}
&\textbf{K}_{\textbf{M}},\textbf{V}_{\textbf{M}}= \mathcal{S}\Big(W_dW_p(\textbf{M})\Big),\\
&\textbf{Q},\textbf{K}_{\textbf{R}},\textbf{V}_{\textbf{R}}= \mathcal{S}\Big(W_dW_p(\textbf{X})\Big),
\\
& \textbf{K}=W_p\Big(\mathcal{C}\big[\textbf{K}_{\textbf{M}}, \textbf{K}_{\textbf{R}}\big]\Big),\\
&\textbf{V}=W_p\Big(\mathcal{C}\big[\textbf{V}_{\textbf{M}}, \textbf{V}_{\textbf{R}}\big]\Big),
\\
& \textbf{A}_{\text{Intra}}=\textbf{V}\otimes \textrm{Softmax$\Big(\textbf{K} \otimes \textbf{Q}/\alpha \Big)$},
\end{split}
%\vspace{-0.95em}
\end{equation}
where $\mathcal{S}(\cdot)$ means the split operation;
$\mathcal{C}[\cdot,\cdot]$ denotes the concatenation operation at channel dimension;
$\alpha$ is a learnable scaling parameter to control the magnitude of the dot product;
$\otimes$ denotes the matrix multiplication~\cite{Zamir2021Restormer};
$\textbf{A}_{\text{Intra}}$ means the output of IntraPPA.

% \noindent \textbf{Inter parser-prompted attention.}
{\flushleft \bf Inter Parser-Prompted Attention.}
The InterPPA, shown in Fig.~\ref{fig: IntraMPA_InterMPA_MPN}(a), re-considers the explicit global fusion between parser features and restoration ones within attention.
Different from IntraPPA, the InterPPA first fuses parser features with restoration ones by the introduced Bidirectional Parser-Prompted Fusion (see Fig.~\ref{fig: Bidirectional Mask-Prompted Fusion}).
Then, the fused features are computed via an attention module.
This process of InterPPA can be formulated as:
\begin{equation}
    \begin{split}
&\textrm{$\textbf{X}_{\text{BiPPF}}$}=\mathcal{P}\Big(\mathbf{X}, \mathbf{M}\Big),
\\
&\textbf{Q}, \textbf{K}, \textbf{V}= \mathcal{S}\Big(W_dW_p{\left(\textrm{$\textbf{X}_{\text{BiPPF}}$}\right)}\Big),
\\
% &\textrm{$\textbf{A}_{\text{Inter}}$}=\mathcal{A}\Big(\textbf{Q}, \textbf{K},\mathbf{V}\Big),  
&\textrm{$\textbf{A}_{\text{Inter}}$}=\textbf{V}\otimes \textrm{Softmax$\Big(\textbf{K} \otimes \textbf{Q}/\alpha \Big)$},
    \end{split}\vspace{-0.5em}
\end{equation}
where $\mathcal{P}(\cdot,\cdot)$ denotes the operation of Bidirectional Parser-Prompted Fusion;
$\textbf{A}_{\text{Intra}}$ means the output of InterPPA.

% \noindent \textbf{Parser-prompted feed-forward network.}
{\flushleft \bf Parser-Prompted Feed-Forward Network.}
Our PPFN is based on GDFN~\cite{Zamir2021Restormer}, which contains two parallel paths with a gating mechanism to control information flow.
We modify the GDFN by integrating parser features into it.
Specifically, we integrate parser features into one of the parallel paths by BiPPF, which allows parser features to effectively guide the restoration process within a pixel-wise modulation perspective.
Then, the integrated features are gated with the features of another path.
The PPFN can be formulated as:
\begin{equation}
    \begin{split}
&\textbf{X}_{1}, \textbf{X}_{2}=\mathcal{S}\Big(W_dW_p\big(\mathbf{X}\big)\Big),
\\
&\textrm{$\textbf{X}_{\text{BiPPF}}$}=\mathcal{P}\big(\textbf{X}_{1}, \mathbf{M}\big),
\\
&\textrm{$\textbf{X}_{\text{PPFN}}$}=W_p\Big(\textrm{$\textbf{X}_{\text{BiPPF}}$} {\odot} \phi(\textbf{X}_{2})\Big) + \mathbf{X},
    \end{split}\vspace{-0.5em}
\end{equation}
where $\phi$ denotes the GELU activation;
$\odot$ means the element-wise multiplication;
$\textrm{$\textbf{X}_{\text{PPFN}}$}$ refers to the output of PPFN.
Following \cite{Zamir2021Restormer}, we use expanding channel capacity factor as 3 to intermediate feature in PPFN.

\subsection{Controllable Parser Feature Propagation}\label{sec: Controllable Mask Feature Propagation }
To enhance the capability of each block in perceiving parser features for improved restoration, we introduce the Controllable Parser Feature Propagation (CPFP). 
CPFP effectively channels valuable parser information into the attention and feed-forward networks within our Parser-Prompted Transformer Block (PPTB). 
In this work, the InterPPA serves as the attention layer in PPTB, while the PPFN is utilized as its feed-forward network.

\subsection{Bidirectional Parser-Prompted Fusion}\label{sec: Bidirectional Mask-Prompted Fusion}
To better fuse parser features with restoration ones, we develop a Bidirectional Parser-Prompted Fusion (BiPPF) module, as shown in Fig.~\ref{fig: Bidirectional Mask-Prompted Fusion}.
Unlike the previous widely-used fusion module SFT~\cite{Wang_2018_CVPR}, we explore the feature fusion manner within bidirectional perspectives to maximize fusion.
Specifically, we first concatenate parser features and restoration ones and then use 1$\times$1 convolution to fuse them.
Then, the fused features are gated with parser features and restoration features by element-wise multiplication, which are respectively added to the original features.
Lastly, we concatenate these added features and use 1$\times$1 convolution to fuse them.
The process can be formulated into:
\begin{equation}\label{eq:gdp}
\begin{split}
& \textrm{$\hat{\textbf{X}}$}=W_p(\textrm{$\textbf{X}$}); \textrm{$\hat{\textbf{M}}$}=W_p(\textrm{$\textbf{M}$}); \textrm{$\textbf{X}_{\text{fusion}}$}=W_p\Big(\mathcal{C}\big[\textrm{$\hat{\textbf{X}}$}, \textrm{$\hat{\textbf{M}}$}\big]\Big),
\\
&\textrm{$\Tilde{\textbf{X}}$}=\textrm{$\textbf{X}_{\text{fusion}}$} {\odot} \textrm{$\hat{\textbf{X}}$} + \textrm{$\hat{\textbf{X}}$};  \textrm{$\Tilde{\textbf{M}}$}=\textrm{$\textbf{X}_{\text{fusion}}$} {\odot} \textrm{$\hat{\textbf{M}}$} + \textrm{$\hat{\textbf{M}}$},
\\
& \textrm{$\textbf{X}_{\text{BiMPF}}$}=W_p\Big(\mathcal{C}\big[\textrm{$\Tilde{\textbf{X}}$}, \textrm{$\Tilde{\textbf{M}}$}\big]\Big) + \textrm{$\textbf{X}$},
\end{split}
\end{equation}
where $\textrm{$\textbf{X}_{\text{BiMPF}}$}$ denotes the output of BiPPF.

\section{Experimental Results}
We evaluate \textbf{PPTformer} on benchmarks for $4$ image restoration tasks: \textbf{(a)} image deraining, \textbf{(b)} single-image defocus deblurring, \textbf{(c)} image desnowing, and \textbf{(d)} low-light image enhancement. 
\begin{table*}[!t]
% \vspace{-4mm}
\begin{center}
% \vspace{-3mm}
\setlength{\tabcolsep}{2pt}
\scalebox{0.855}{
\begin{tabular}{l | c | c  c  c cccccccccccc}
\shline
 % &  &  &  &  \\
 Benchmark & \textbf{Metrics}&DerainNet&SEMI &UMRL&RESCAN &PreNet &MSPFN&DCSFN  & MPRNet& SPAIR&Uformer &MAXIM-2S&SFNet&\textbf{PPTformer}\\
\shline
\multirow{2}{*}{\textbf{Test100}}& PSNR~$\uparrow$ &22.77  &22.35  &24.41&25.00&24.81&27.50&27.46&30.27&30.35&29.17&31.17&31.47& 31.48\\
& SSIM~$\uparrow$ & 0.810 &0.788&0.829&0.835&0.851&0.876&0.887&0.897&0.909&0.880&0.922& 0.919& 0.922\\
\shline
\multirow{2}{*}{\textbf{Rain100H}}& PSNR~$\uparrow$     & 14.92&16.56&26.01&26.36&26.77&28.66&28.98&30.41&30.95&30.06&30.81&31.90&31.77 \\
& SSIM~$\uparrow$     &0.592&0.486&0.832&0.786 &0.858&0.860&0.887&0.890&0.892&0.884&0.903&0.908& 0.907 \\
\shline
\multirow{2}{*}{\textbf{Test2800}}& PSNR~$\uparrow$    &24.31&24.43&29.97&31.29&31.75&32.82&30.96&33.64& 33.34&33.36&33.80&33.69&34.01\\
& SSIM~$\uparrow$     &0.861&0.782&0.905&0.904&0.916&0.930&0.903&0.938&0.936&0.935&0.943&0.937&0.945 \\
\shline
\multirow{2}{*}{\textbf{\textbf{Average}}}& PSNR~$\uparrow$   &20.67&21.11&26.80&27.55&27.78&29.66&29.13&31.44&31.55& 30.86&31.93&\textit{32.35}&\textbf{32.42}\\

& SSIM~$\uparrow$  &0.754&0.685&0.855&0.842&0.875&0.889&0.892&0.908&0.912& 0.900&\textit{0.923}&0.921&\textbf{0.925}\\
\shline
\end{tabular}}
% \vspace{-3mm}
\caption{Image deraining results. Our PPTformer advances recent 13 state-of-the-art approaches on average.}
\label{table: deraining}
\end{center}
% \vspace{-5mm}
\end{table*}
\begin{figure*}[!t]
% \footnotesize
\centering
\begin{center}
\begin{tabular}{ccccccccc}
\includegraphics[width=0.1205\linewidth]{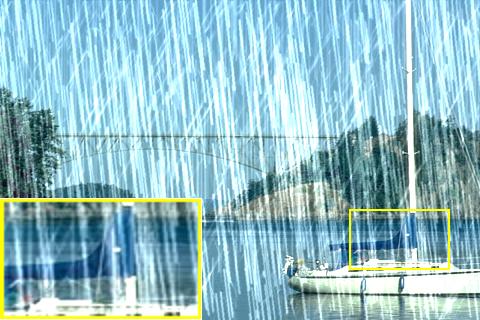} &\hspace{-4.5mm}
\includegraphics[width=0.1205\linewidth]{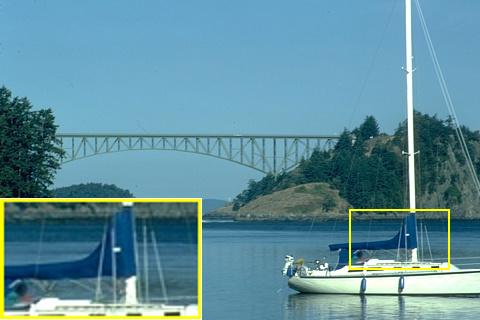} &\hspace{-4.5mm}
\includegraphics[width=0.1205\linewidth]{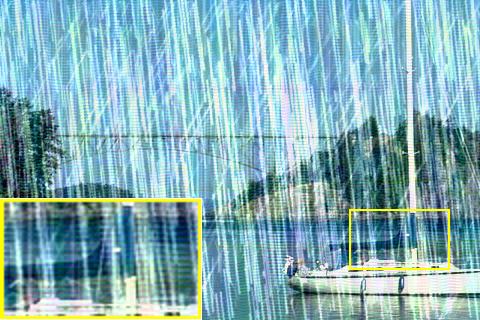} &\hspace{-4.5mm}
\includegraphics[width=0.1205\linewidth]{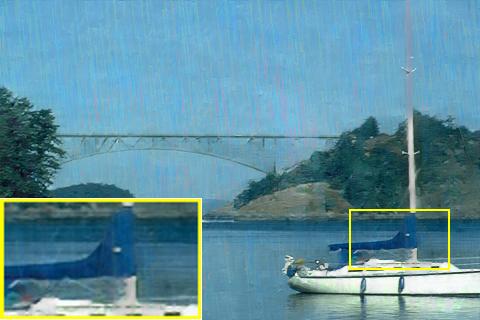} &\hspace{-4.5mm}
\includegraphics[width=0.1205\linewidth]{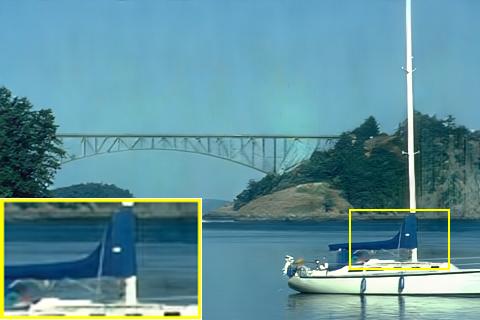} &\hspace{-4.5mm}
\includegraphics[width=0.1205\linewidth]{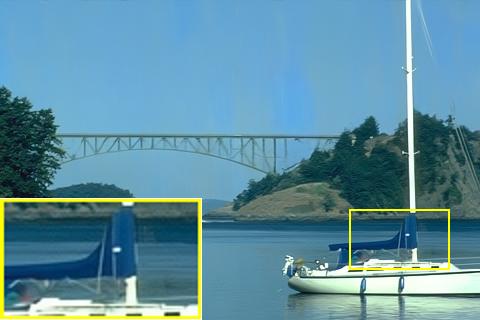} &\hspace{-4.5mm}
\includegraphics[width=0.1205\linewidth]{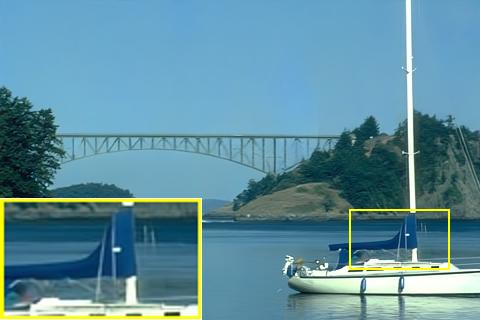} &\hspace{-4.5mm}
\includegraphics[width=0.1205\linewidth]{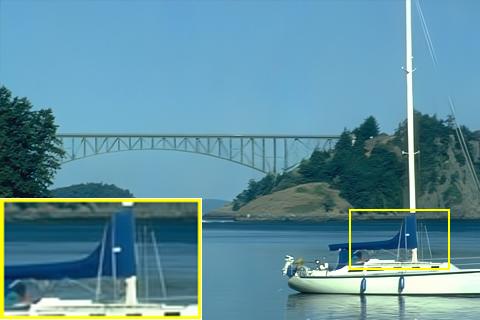}
\\
Input&\hspace{-4.5mm}  GT&\hspace{-4.5mm}  SEMI&\hspace{-4.5mm}   RESCAN&\hspace{-4.5mm}  DCSFN&\hspace{-4.5mm}  MAXIM&\hspace{-4.5mm}  SFNet &\hspace{-4.5mm}  \textbf{Ours} 
\end{tabular}
% \vspace{-3mm}
\caption{Image deraining example on Rain100H~\cite{derain_jorder_yang}.
Our PPTformer recovers results with finer structures.
%
% Best viewed with zoom-in.
}
\label{fig:Image deraining example.}
\end{center}
% \vspace{-5mm}
\end{figure*}
\subsection{Implementation Details}
We train PPTformer using the AdamW optimizer~\cite{adamw} with the initial learning rate $5e^{-4}$ that is gradually reduced to $1e^{-7}$ with the cosine annealing~\cite{loshchilov2016sgdr}.
The training patch size is set as $256$$\times$$256$ pixels.
For downsampling and upsampling, we adopt pixel-unshuffle and pixel-shuffle~\cite{shi2016real_pixelshuffle}, respectively. 
For the parser, we first employ the SAM~\cite{kirillov2023segany} to generate the parser before training the PPTformer.
Then, we treat the parser as an image to input the PPTformer for efficiency as loading the SAM consumes much more memory when training.
To constrain the training of PPTformer, we use the same loss function~\cite{Kong_2023_CVPR_fftformer} with default parameters.
% More details on datasets and additional visual results are presented in the supplementary material. 

%
\subsection{Main Results}

{\flushleft \bf Image Deraining Results.}
Similar to existing methods~\cite{mspfn2020,Zamir_2021_CVPR_mprnet,purohit2021spatially_spair}, we report PSNR/SSIM scores using Y channel in YCbCr color by comparing with DerainNet~\cite{fu2017clearing}, SEMI~\cite{wei2019semi}, DIDMDN~\cite{zhang2018density}, UMRL~\cite{yasarla2019uncertainty}, RESCAN~\cite{li2018recurrent}, PreNet~\cite{ren2019progressive}, MSPFN~\cite{mspfn2020},DCSFN~\cite{dcsfn-wang-mm20}, MPRNet~\cite{Zamir_2021_CVPR_mprnet}, SPAIR~\cite{purohit2021spatially_spair}, Uformer~\cite{wang2021uformer}, MAXIM-2S~\cite{maxim_Tu_2022_CVPR}, SFNet~\cite{cui2023selective}.
Tab.~\ref{table: deraining} shows that our PPTformer outperforms current state-of-the-art approaches when averaged across these three datasets, including Test100~\cite{zhang2019image}, Rain100H~\cite{yang2017deep}, and Test2800~\cite{fu2017clearing}. 
Compared to recent the best method SFNet~\cite{cui2023selective}, our PPTformer achieves $0.07$~dB improvement on average. 
On individual datasets, the gain can be as large as $0.32$~dB on Test2800~\cite{fu2017removing}.
Fig.~\ref{fig:Image deraining example.} further presents a challenging example on Rain100H~\cite{yang2017deep}, where our PPTformer generates a clearer image with finer details.

\begin{table*}[!t]
\begin{center}
% \vspace{-3mm}
\setlength{\tabcolsep}{1.75pt}
\scalebox{0.8675}{
\begin{tabular}{l | c c c c | c c c c | c c c c }
\shline
%\rowcolor{color3} 
\multirow{2}{*}{\textbf{Method}}  & \multicolumn{4}{c|}{\textbf{Indoor Scenes}} & \multicolumn{4}{c|}{\textbf{Outdoor Scenes}} & \multicolumn{4}{c}{\textbf{Combined}} \\
\cline{2-13}
%\rowcolor{color3} 
\multicolumn{1}{c|}{}    & PSNR~$\uparrow$ & SSIM~$\uparrow$& MAE~$\downarrow$ & LPIPS~$\downarrow$  & PSNR~$\uparrow$ & SSIM~$\uparrow$& MAE~$\downarrow$ & LPIPS~$\downarrow$  & PSNR~$\uparrow$ & SSIM~$\uparrow$& MAE~$\downarrow$ & LPIPS~$\downarrow$   \\
\shline
EBDB~\cite{karaali2017edge_EBDB} & 25.77 & 0.772 & 0.040 & 0.297 & 21.25 & 0.599 & 0.058 & 0.373 & 23.45 & 0.683 & 0.049 & 0.336 \\
DMENet~\cite{lee2019deep_dmenet}  & 25.50 & 0.788 & 0.038 & 0.298 & 21.43 & 0.644 & 0.063 & 0.397 & 23.41 & 0.714 & 0.051 & 0.349 \\
JNB~\cite{shi2015just_jnb} & 26.73 & 0.828 & 0.031 & 0.273 & 21.10 & 0.608 & 0.064 & 0.355 & 23.84 & 0.715 & 0.048 & 0.315 \\
DPDNet~\cite{abdullah2020dpdd} &26.54 & 0.816 & 0.031 & 0.239 & 22.25 & 0.682 & 0.056 & 0.313 & 24.34 & 0.747 & 0.044 & 0.277\\
% KPAC$_S$~\cite{son2021single_kpac} & \qmarks  &  \qmarks  &  \qmarks &  \qmarks  &  \qmarks  &  \qmarks &  \qmarks  &  \qmarks  & 25.24 & 0.845 & ???  & 0.225\\
KPAC~\cite{son2021single_kpac} & 27.97 & 0.852 & 0.026 & 0.182 & 22.62 & 0.701 & 0.053 & 0.269 & 25.22 & 0.774 & 0.040 & 0.227 \\
IFAN~\cite{Lee_2021_CVPRifan} & 28.11 & 0.861 & 0.026 & 0.179 & 22.76  & 0.720 & 0.052  & 0.254  & 25.37 & 0.789& 0.039 & 0.217\\
Restormer~\cite{Zamir2021Restormer}& \textit{28.87}  & \textbf{0.882}  & \textit{0.025} &\textbf{0.145} & \textit{23.24} &\textbf{0.743}  & \textit{0.050} & \textbf{0.209}& 25.98  &\textbf{0.811} & \textit{0.038} & \textbf{0.178}   \\

% SFNet~\cite{cui2023selective}& 29.16  &0.878  & 0.023 & 0.168 & 23.45  & 0.747  & 0.049 & 0.244  & 26.23  & 0.811  & 0.037  & 0.207  \\
\shline
\textbf{PPTformer}&\textbf{29.16}  &\textit{0.880 } &\textbf{0.024} &\textit{0.156}  & \textbf{23.26}  &\textit{ 0.737}  &\textbf{0.050} & \textit{0.227}  & \textbf{26.13}  &\textit{0.807}  & \textbf{0.037} &  \textit{0.193} \\
\shline
\end{tabular}}
% \vspace{-3mm}
\caption{Single-image defocus deblurring comparisons on the DPDD test set~\cite{abdullah2020dpdd} (containing 37 indoor and 39 outdoor scenes). 
% \textbf{S:} single-image defocus deblurring. \textbf{D:} dual-pixel defocus deblurring. 
Our PPTformer achieves the best PSNR. 
%
% Best viewed with zoom-in.
}
\label{table:dpdeblurring}
\end{center}
% \vspace{-5mm}
\end{table*}
\begin{figure*}[!t]
\centering
\begin{center}
\begin{tabular}{ccccccccc}
\includegraphics[width=0.1205\linewidth]{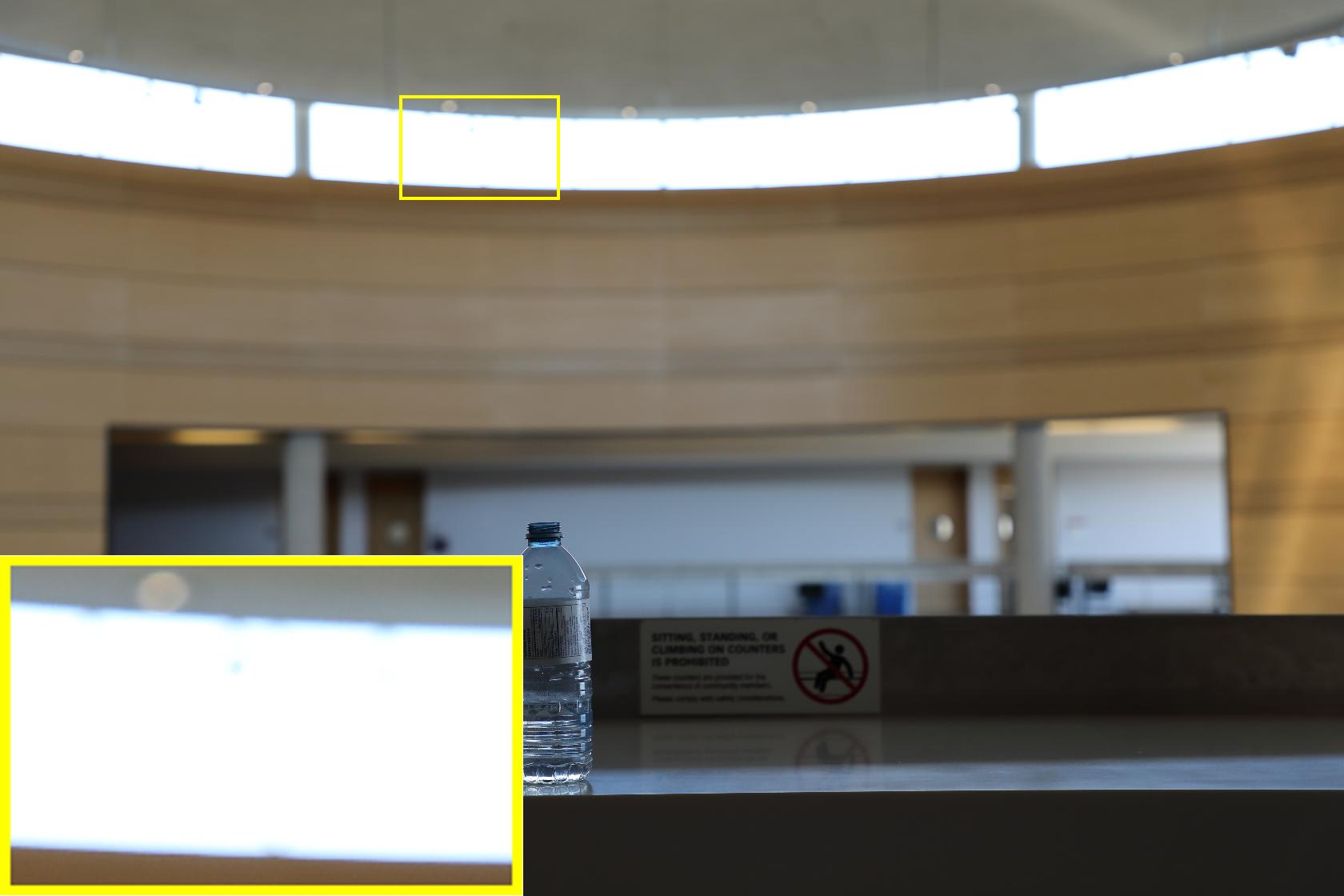} &\hspace{-4.5mm}
\includegraphics[width=0.1205\linewidth]{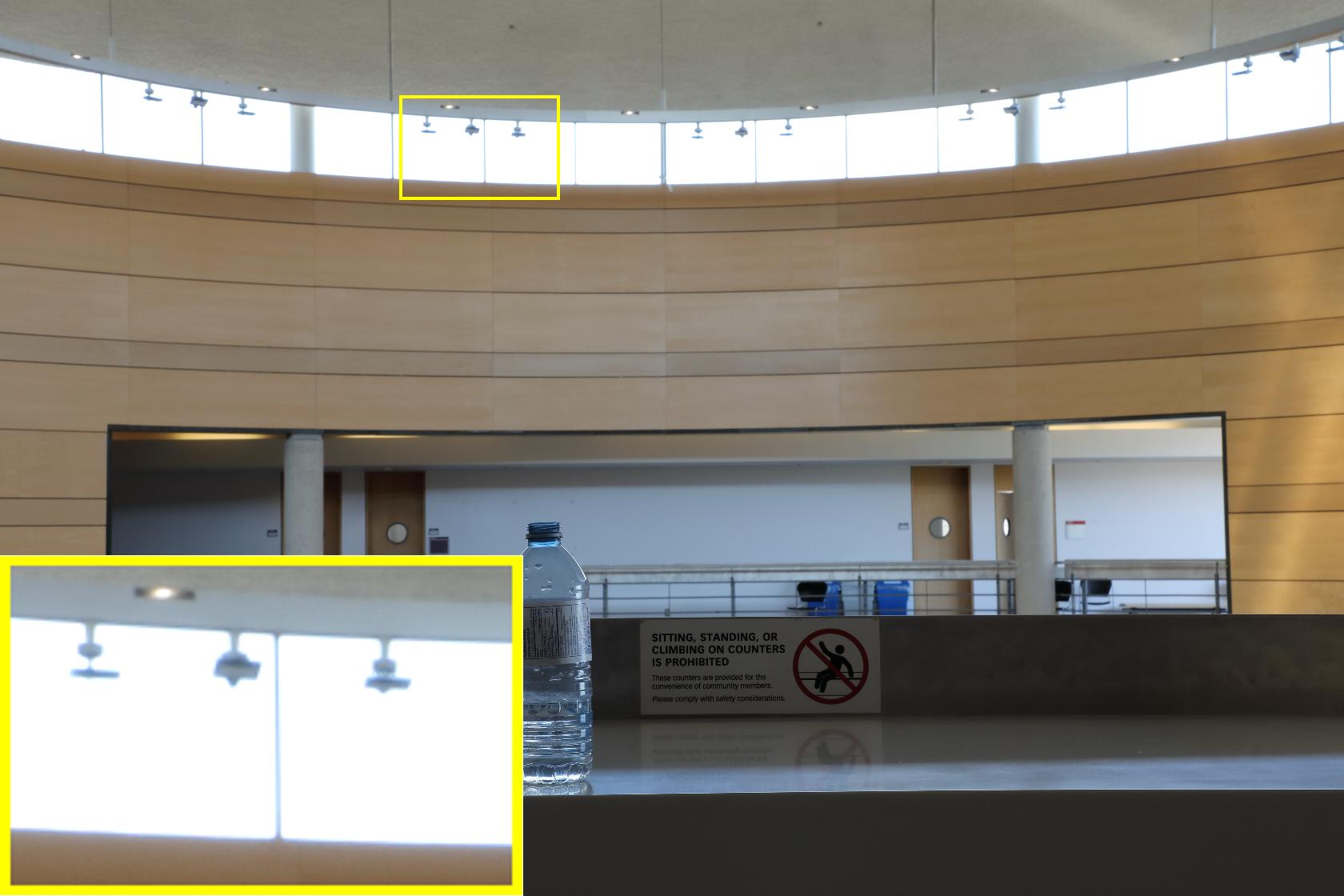} &\hspace{-4.5mm}
\includegraphics[width=0.1205\linewidth]{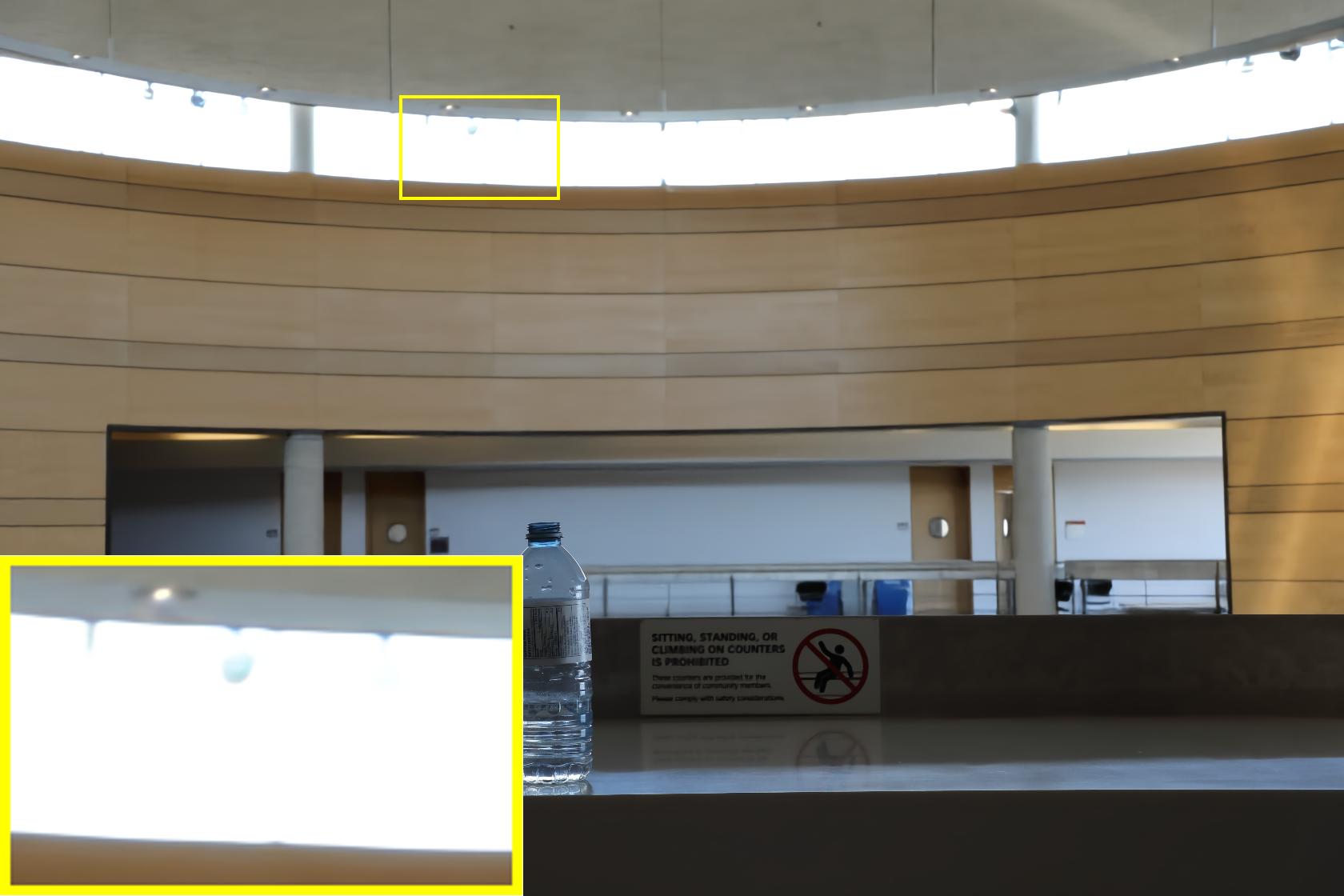} &\hspace{-4.5mm}
\includegraphics[width=0.1205\linewidth]{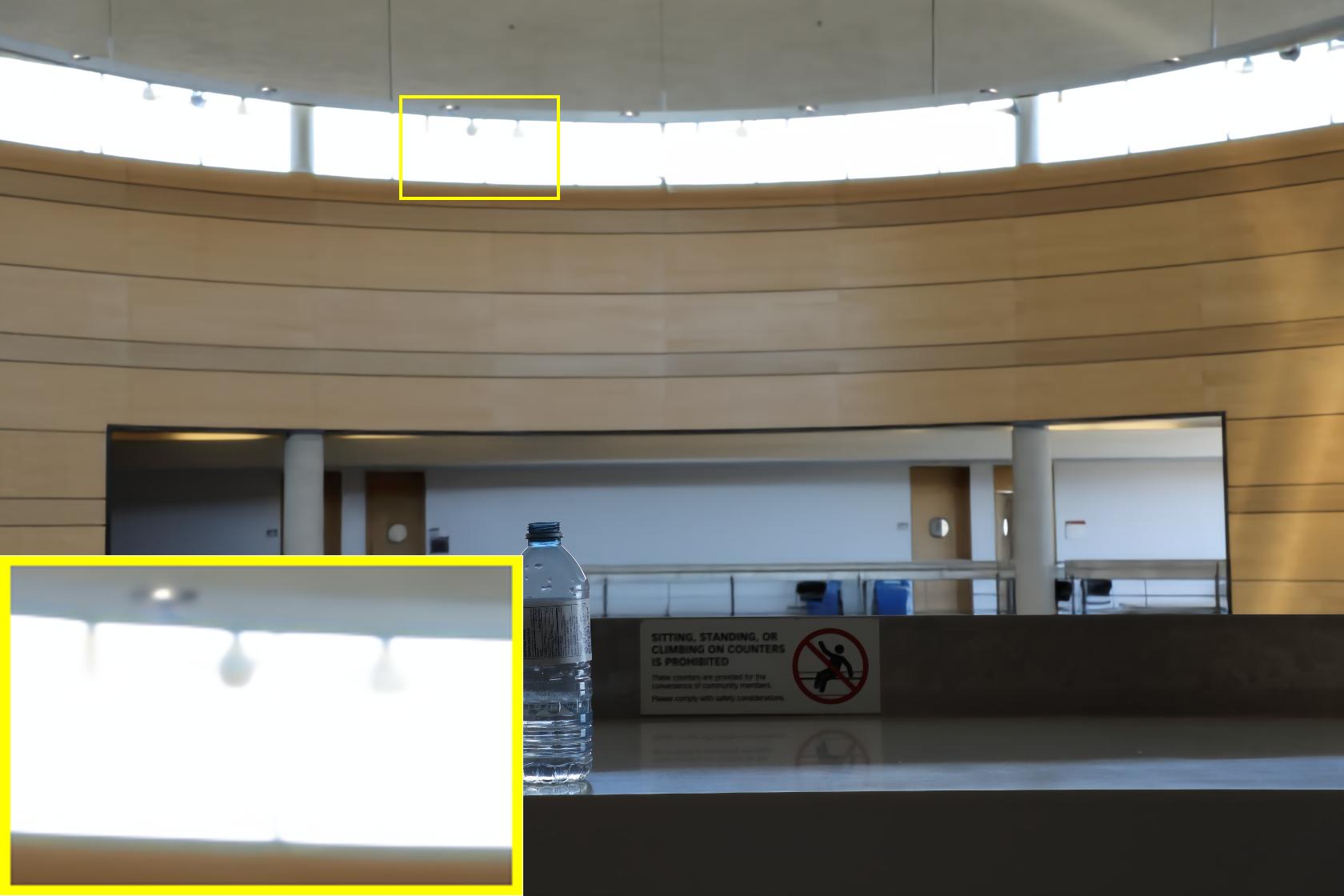} &\hspace{-4.5mm}
\includegraphics[width=0.1205\linewidth]{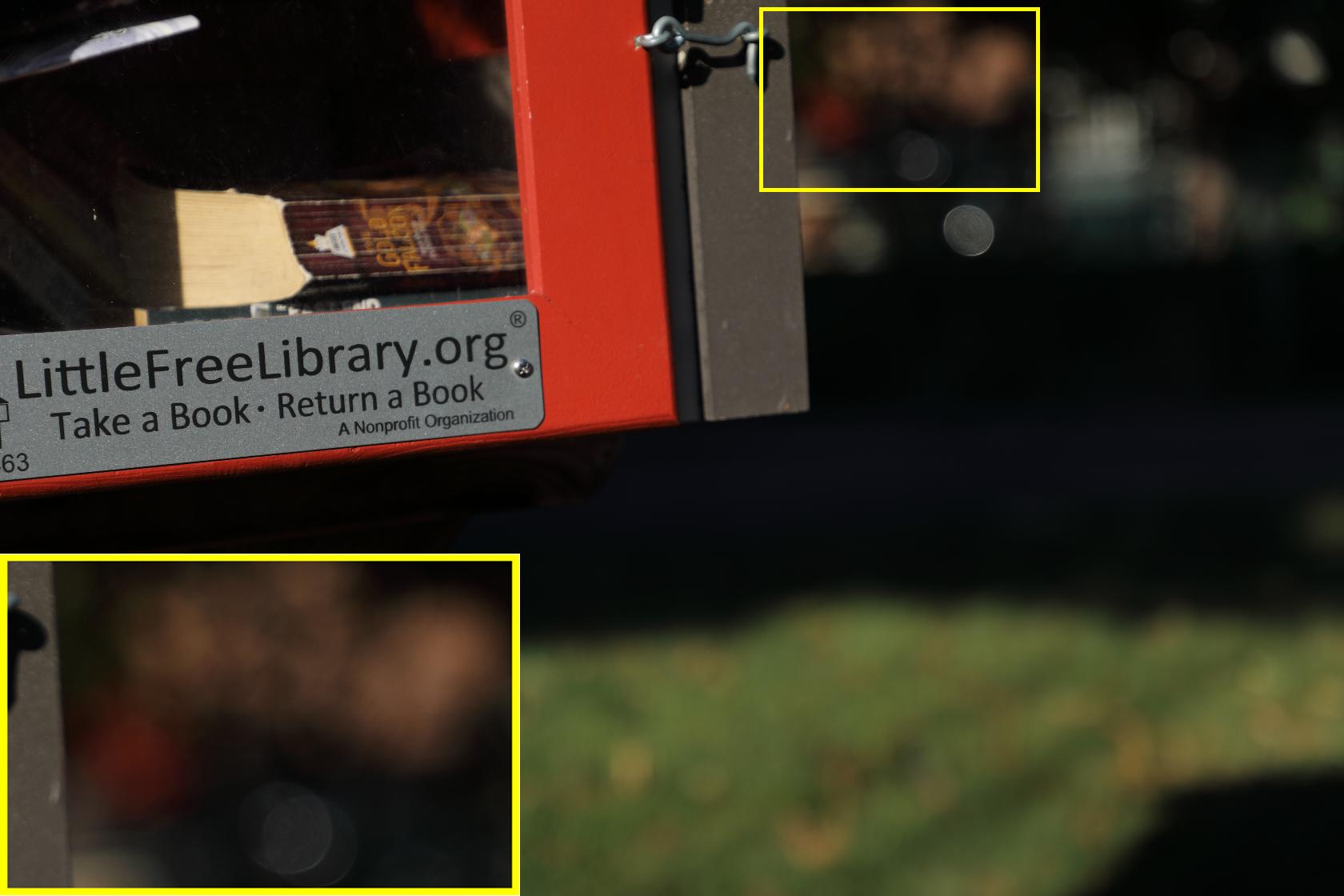} &\hspace{-4.5mm}
\includegraphics[width=0.1205\linewidth]{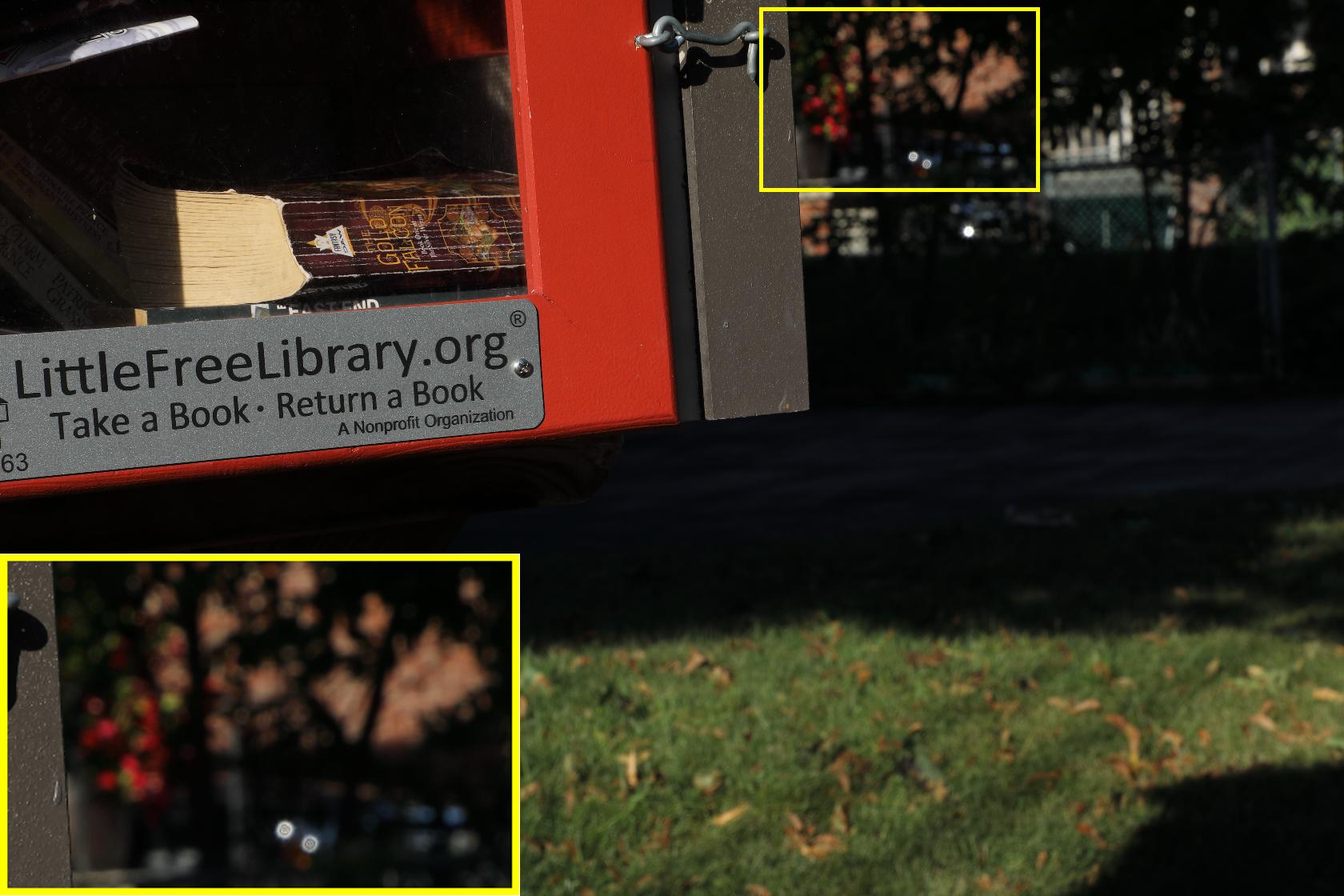} &\hspace{-4.5mm}
\includegraphics[width=0.1205\linewidth]{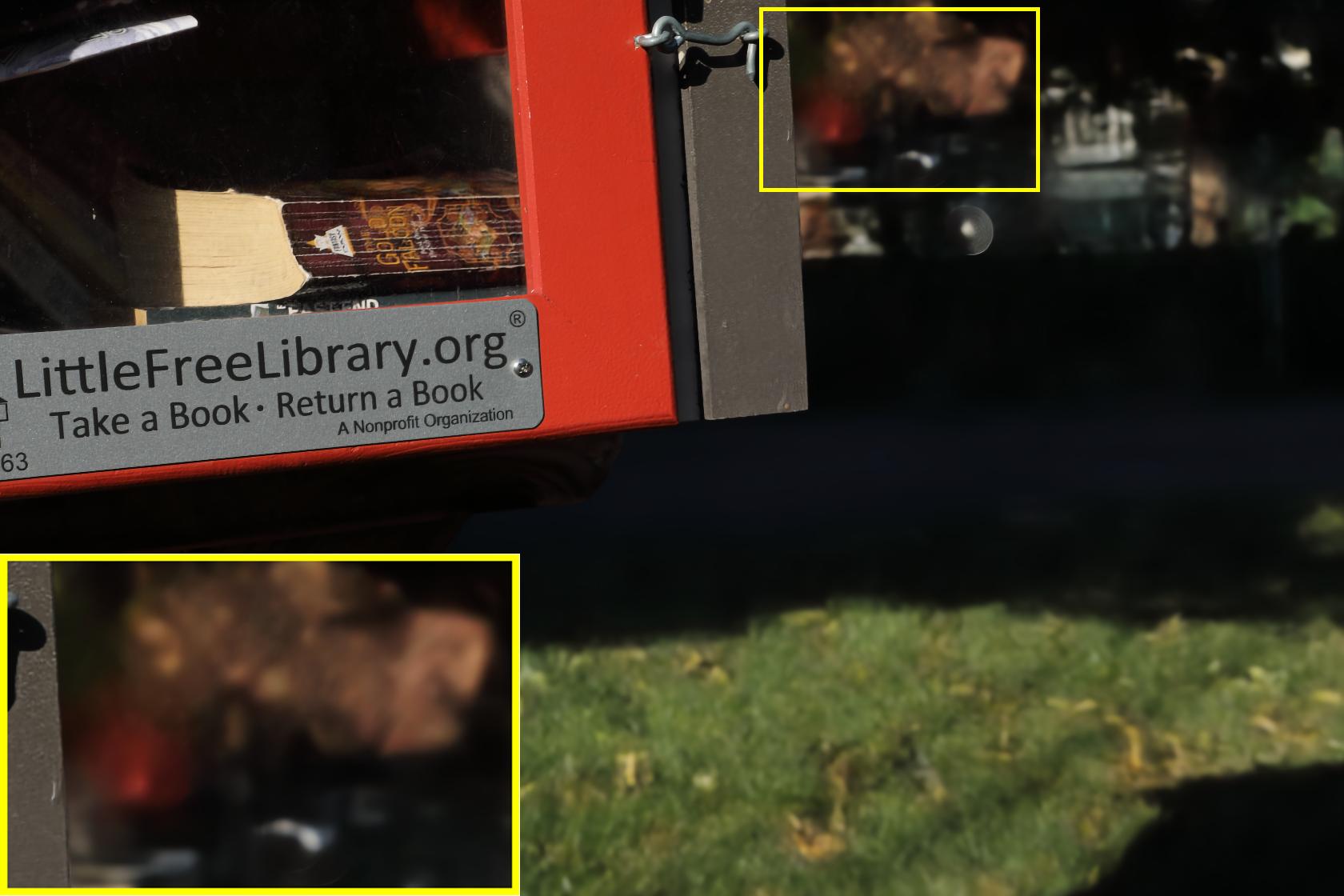} &\hspace{-4.5mm}
\includegraphics[width=0.1205\linewidth]{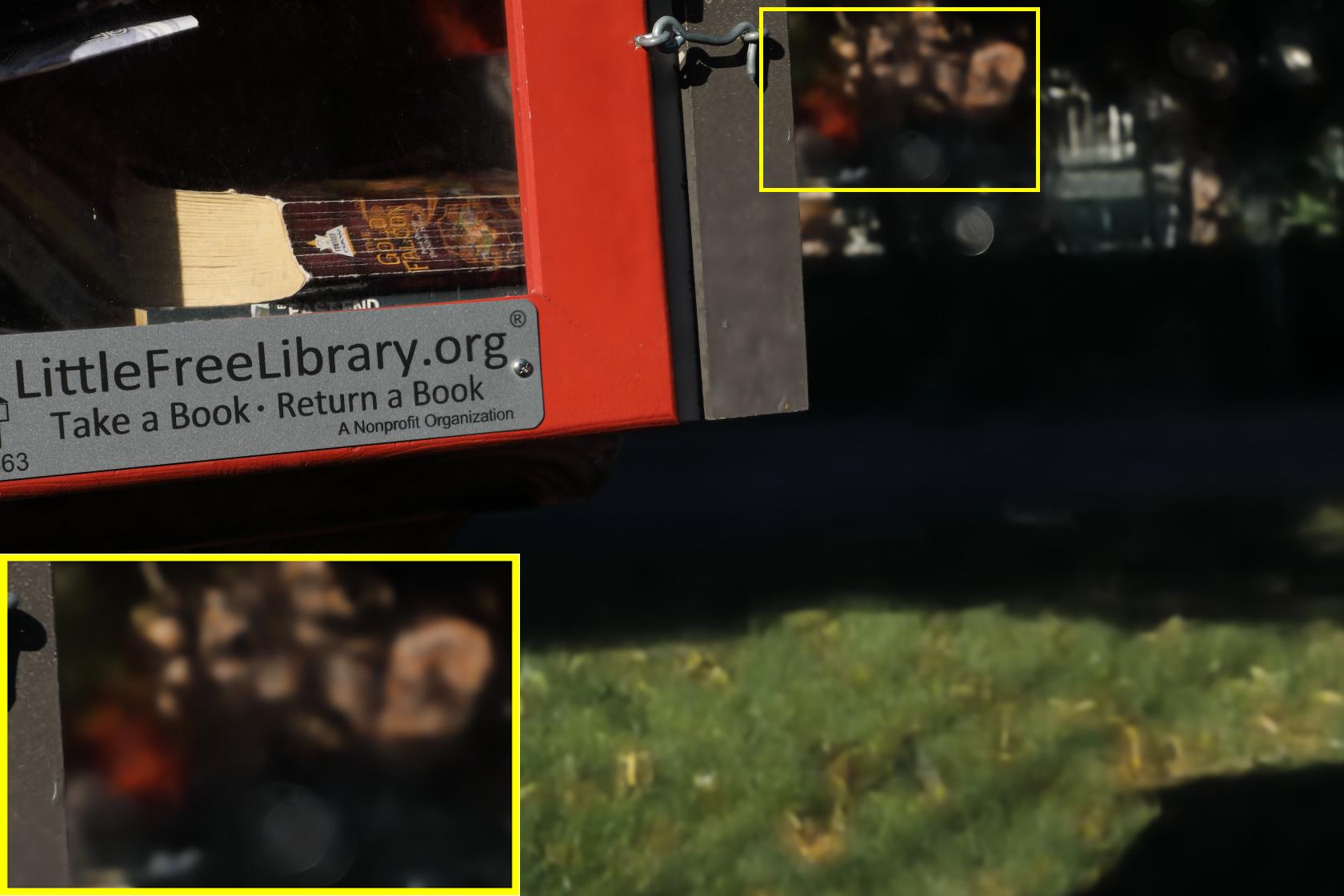}
\\
Input&\hspace{-4.5mm}  GT&\hspace{-4.5mm} Restormer&\hspace{-4.5mm}  \textbf{Ours}&\hspace{-4.5mm} Input&\hspace{-4.5mm}  GT&\hspace{-4.5mm}  Restormer&\hspace{-4.5mm}  \textbf{Ours} 
\end{tabular}
% \vspace{-3mm}
\caption{Single-image defocus deblurring example in indoor and outdoor scenes on DPDD test set~\cite{abdullah2020dpdd}.
Our PPTformer is able to recover clearer results with sharper structures.
Best viewed with zoom-in.
}
\label{fig:Single-image defocus deblurring example.}
\end{center}
% \vspace{-5mm}
\end{figure*}

{\flushleft \bf Single-Image Defocus Deblurring Results.}
Tab.~\ref{table:dpdeblurring} summarises the single-image defocus deblurring results compared with conventional defocus deblurring methods (EBDB~\cite{karaali2017edge_EBDB} and JNB~\cite{shi2015just_jnb}) as well as learning-based approaches~\cite{abdullah2020dpdd,son2021single_kpac,Lee_2021_CVPRifan,Zamir2021Restormer} on the DPDD dataset~\cite{abdullah2020dpdd}. 
As one can see our PPTformer achieves comparable results with Restormer~\cite{Zamir2021Restormer} but outperforms other state-of-the-art approaches in terms of PSNR/SSIM/MAE/LPIPS metrics. 
In particular, PPTformer yields the best PSNR and MAE in all categories.
Compared with Restormer~\cite{Zamir2021Restormer}, our method improves 0.15dB PSNR gains on combined scenes. 
Fig.~\ref{fig:Single-image defocus deblurring example.} presents two visual examples of indoor and outdoor scenes, where our PPTformer is more effective in recovering image structures from complex blurry scenes than Restormer.

\begin{figure*}[!t]
% \vspace{-1mm}
\begin{center}
\begin{tabular}{cccccccccc}
\includegraphics[width = 0.1375\linewidth]{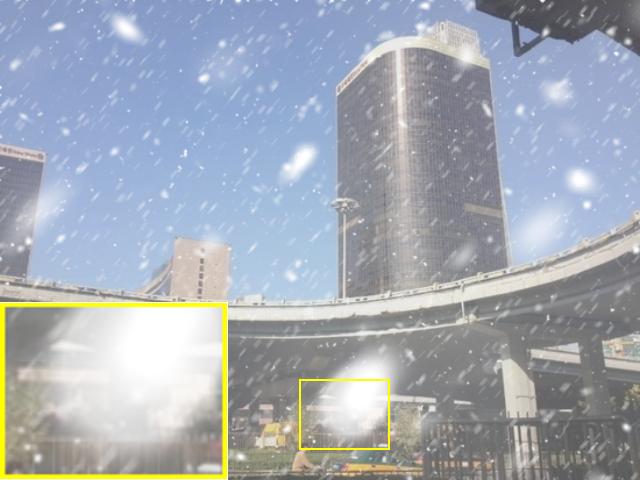}&\hspace{-4.5mm}
\includegraphics[width = 0.1375\linewidth]{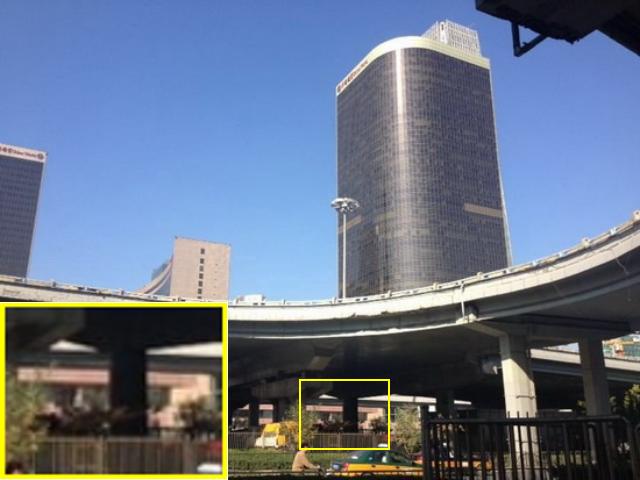}&\hspace{-4.5mm}
\includegraphics[width = 0.1375\linewidth]{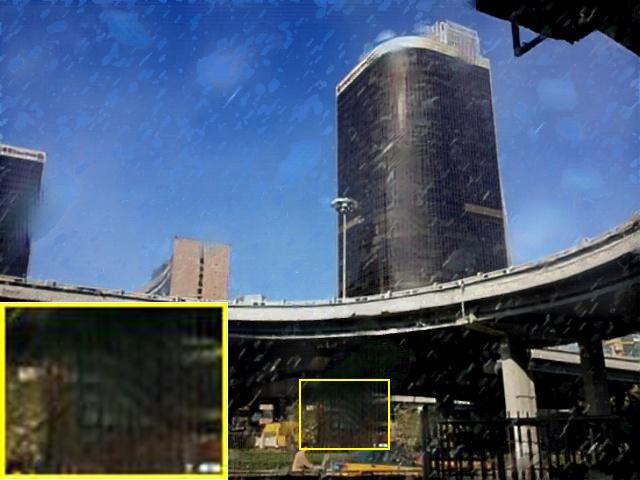}&\hspace{-4.5mm}
\includegraphics[width = 0.1375\linewidth]{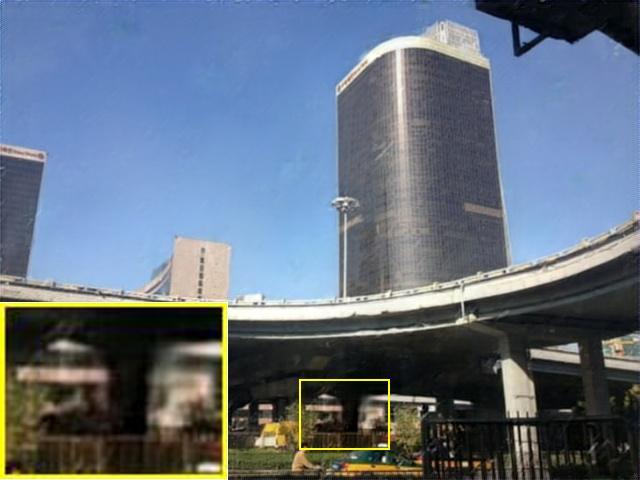}&\hspace{-4.5mm}
\includegraphics[width = 0.1375\linewidth]{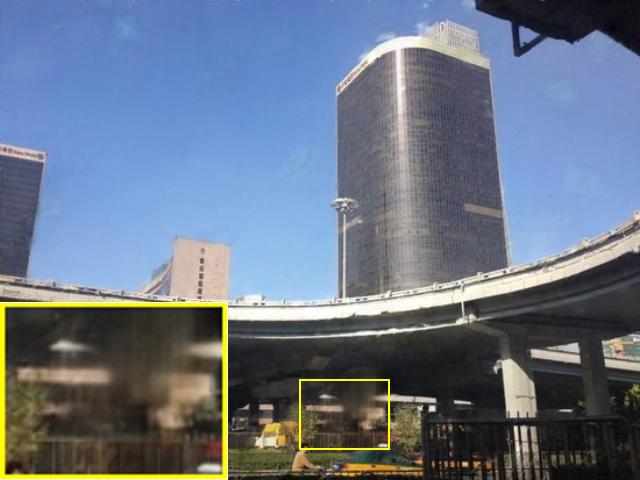}&\hspace{-4.5mm}
\includegraphics[width = 0.1375\linewidth]{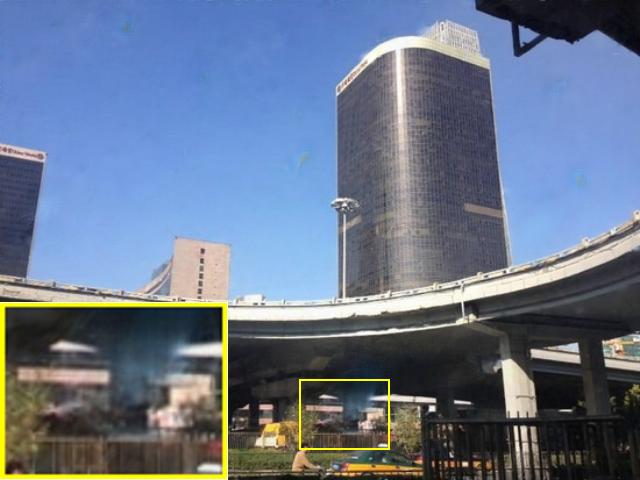}&\hspace{-4.5mm}
\includegraphics[width = 0.1375\linewidth]{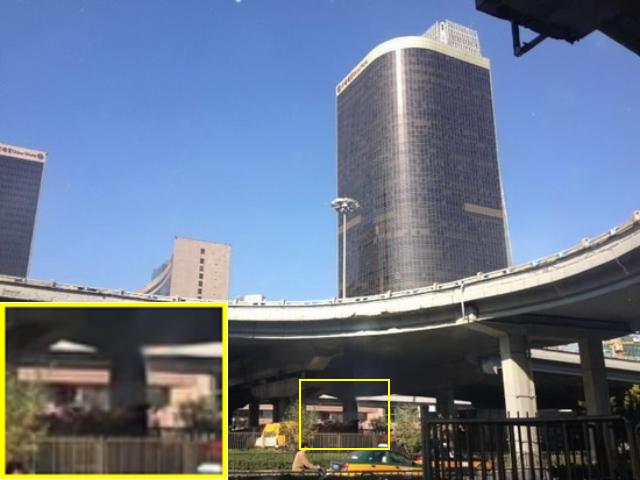}
\\
Input &\hspace{-4.5mm}   GT   &\hspace{-4.5mm}   JSTASR  &\hspace{-4.5mm} HDCWNet&\hspace{-4.5mm}  Uformer&\hspace{-4.5mm} Restormer &\hspace{-4.5mm} \textbf{Ours}
\\
\end{tabular}
% \vspace{-3mm}
\caption{ Image desnowing example on CSD (2000)~\cite{chen2021all}.
Our PPTformer recovers cleaner results.
%
% Best viewed with zoom-in.
%
}
\label{fig: Visual comparisons of image desnowing on CSD (2000).}
\end{center}
% \vspace{-5mm}
\end{figure*}

{\flushleft \bf Image Desnowing Results.}
For the image desnowing task, we compare our PPTformer on the CSD~\cite{chen2021all}, SRRS~\cite{chen2020jstasr}, and Snow100K~\cite{liu2018desnownet} datasets with existing state-of-the-art competitors~\cite{liu2018desnownet,chen2020jstasr,chen2021all,chen2022msp,valanarasu2022transweather}.
Except for the above methods with specific designs for image desnowing, we also compare with recent Transformer-based general image restoration approaches like Restormer~\cite{Zamir2021Restormer} and Uformer~\cite{wang2021uformer}.
Tab.~\ref{table:Image desnowing} shows that our PPTformer respectively yields a $0.84$~dB, $1.25$~dB, and $1.47$~dB PSNR improvement over the state-of-the-art approach~\cite{Zamir2021Restormer} on the CSD~\cite{chen2021all}, SRRS~\cite{chen2020jstasr}, and Snow100K~\cite{liu2018desnownet} benchmarks. 
The visual example in Fig.~\ref{fig: Visual comparisons of image desnowing on CSD (2000).} demonstrates that our PPTformer is able to remove spatially varying snowflakes than state-of-the-art approaches.

\begin{figure*}[!t]
% \vspace{-1mm}
\begin{center}
\begin{tabular}{cccccccccc}
\includegraphics[width = 0.1375\linewidth]{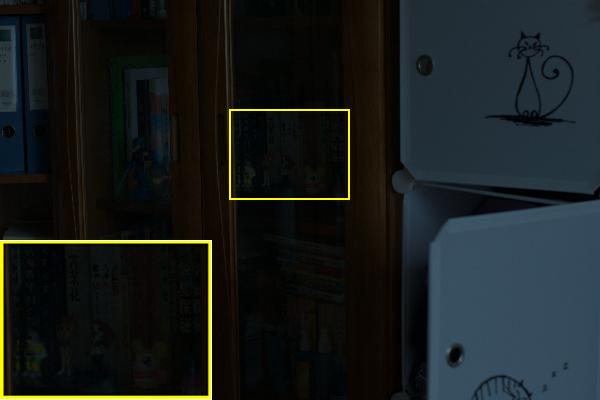}&\hspace{-4.5mm}
\includegraphics[width = 0.1375\linewidth]{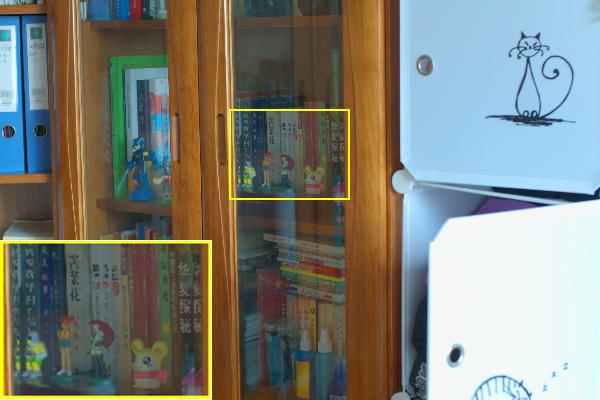}&\hspace{-4.5mm}
\includegraphics[width = 0.1375\linewidth]{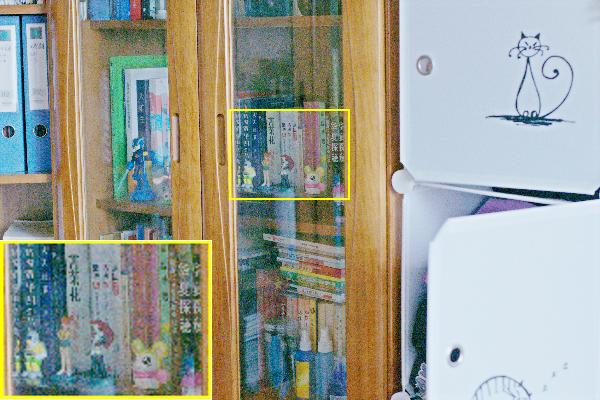}&\hspace{-4.5mm}
\includegraphics[width = 0.1375\linewidth]{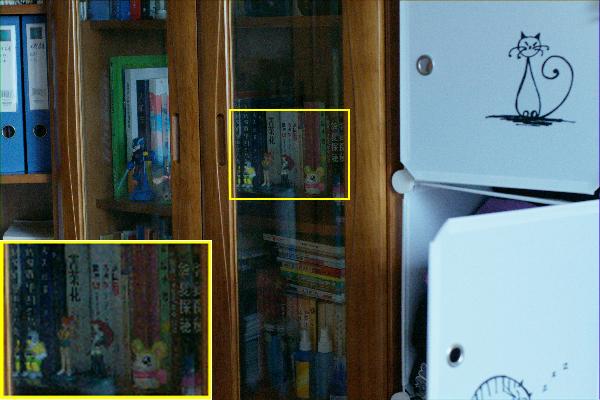}&\hspace{-4.5mm}
\includegraphics[width = 0.1375\linewidth]{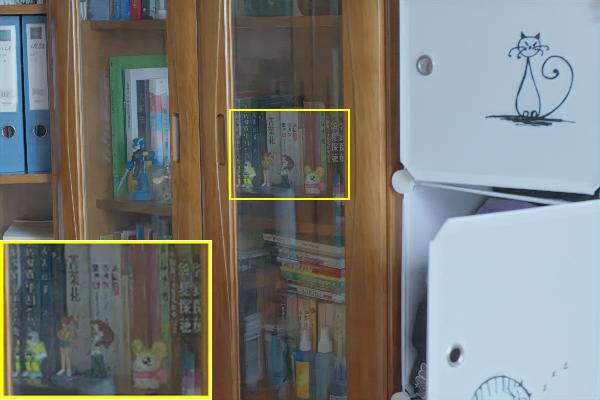}&\hspace{-4.5mm}
\includegraphics[width = 0.1375\linewidth]{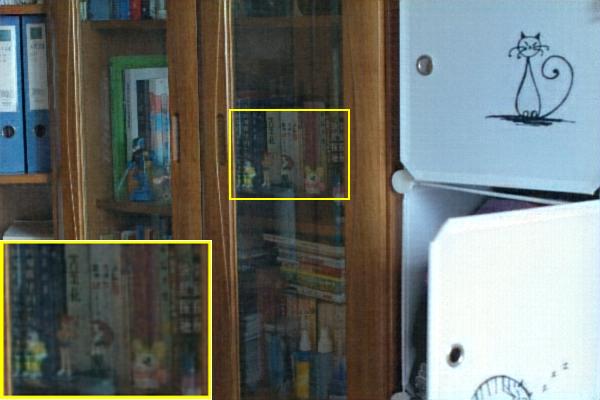}&\hspace{-4.5mm}
\includegraphics[width = 0.1375\linewidth]{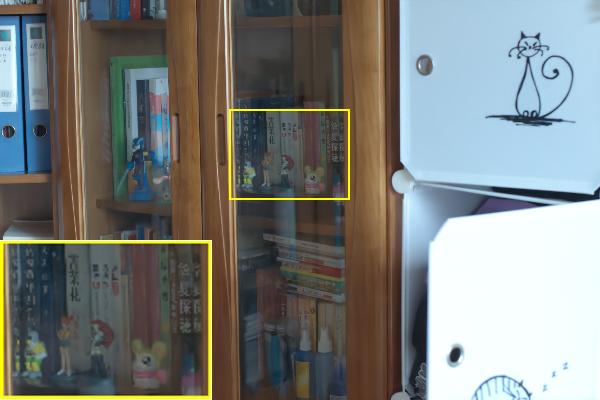}
\\
 Input  &\hspace{-4.5mm}  GT  &\hspace{-4.5mm}   ZeroDCE &\hspace{-4.5mm}  SCI&\hspace{-4.5mm} URetinex&\hspace{-4.5mm} UHDFour  &\hspace{-4.5mm} \textbf{Ours}
\\
\end{tabular}
% \vspace{-3mm}
\caption{Low-light image enhancement example on LOL~\cite{retinexnet_wei_bmvc18}.
Our PPTformer restores more vivid colors.
%
% Best viewed with zoom-in.
}
\label{fig: Low-light image enhancement example on LOL.}
\end{center}
% \vspace{-5mm}
\end{figure*}

{\flushleft \bf Low-Light Image Enhancement Results.}
We perform the low-light image enhancement experiment on both LOL~\cite{retinexnet_wei_bmvc18} and LOL-v2~\cite{lol-v2}. 
Tab.~\ref{table:Low-light image enhancement} summarizes the quantitative results, compared with Retinex-Net~\cite{retinexnet_wei_bmvc18}, Zero-DCE~\cite{zerodce_lowlight_guo}, AGLLNet~\cite{AGLLNet}, Zhao~et al.~\cite{zhao_lie}, RUAS~\cite{RUAS_liu_cvpr21}, SCI~\cite{ma2022toward}, URetinex-Net~\cite{Wu_2022_CVPR}, UHDFour~\cite{Li2023ICLR_uhdfour}.
Compared to recent works UHDFour~\cite{Li2023ICLR_uhdfour}, our method receives $2.39$~dB and $2.74$~dB PSNR gains on LOL~\cite{retinexnet_wei_bmvc18} and LOL-v2~\cite{lol-v2}, respectively. 
Fig.~\ref{fig: Low-light image enhancement example on LOL.} shows our PPTformer is able to generate clearer results with more vivid colors.
\begin{table}[!t]
\begin{center}
\centering
% \vspace{-3mm}
\setlength{\tabcolsep}{2.4pt}
\scalebox{0.825}{
\begin{tabular}{l | c  c | c  c | cccccccccccc}
\shline
\multirow{2}{*}{\textbf{Method}} & \multicolumn{2}{c|}{\textbf{CSD (2000)}}& \multicolumn{2}{c|}{\textbf{SRRS (2000)}}& \multicolumn{2}{c}{\textbf{Snow100K (2000)}}
 \\
  & PSNR~$\uparrow$ & SSIM~$\uparrow$ & PSNR~$\uparrow$ &SSIM~$\uparrow$ & PSNR~$\uparrow$ & SSIM~$\uparrow$\\
\shline
DesnowNet& 20.13&0.81  & 20.38&0.84 &  30.50&0.94 \\
STASR&  27.96&0.88  & 25.82&0.89&23.12&0.86 \\
HDCW-Net&  29.06&0.91  & 27.78&0.92&31.54&0.95 \\
TransWeather&   31.76&0.93   &  28.29&0.92& 31.82&0.95 \\
MSP-Former&  33.75&0.96 &  30.76&0.95 & 33.43&0.96  \\
Uformer&33.80&0.96 & 30.12&0.96 & 33.81&0.94 \\
Restormer&  35.43&0.97 &  32.24&0.96  & 34.67&0.95\\
\shline
\textbf{PPTformer} &\textbf{36.27} &\textbf{0.99}&\textbf{33.49}&\textbf{0.98}&\textbf{36.14}&\textbf{0.96}  \\
\shline
\end{tabular}
}
% \vspace{-3mm}
\caption{Image desnowing results on CSD (2000)~\cite{chen2021all}, SRRS (2000)~\cite{chen2020jstasr}, and Snow100K (2000)~\cite{liu2018desnownet}.
% Our \textbf{SamRestorer} achieves the best metrics on the image desnowing.
} 
\label{table:Image desnowing}
\end{center}
% \vspace{-5mm}
\end{table}

\begin{table}[!t]
\begin{center}
\centering
% \vspace{-3mm}
\setlength{\tabcolsep}{5.1pt}
\scalebox{0.99999}{
\begin{tabular}{l|cc|cc}
\shline
\multirow{2}{*}{\textbf{Method}}&\multicolumn{2}{c|}{ \textbf{LOL}}& \multicolumn{2}{c}{ \textbf{LOL-v2}}
\\
&PSNR~$\uparrow$& SSIM~$\uparrow$
  &PSNR~$\uparrow$ & SSIM~$\uparrow$
\\
\shline
Retinex-Net&16.77 &0.54 &15.43&0.64\\
Zero-DCE&16.79&0.67  &12.84&0.54 \\
AGLLNet&17.52&0.77  &20.69&0.78 \\
Zhao~et al.&21.67 & 0.87 &18.84&0.78 \\
RUAS&16.44 &0.70  &15.48&0.67 \\
SCI&14.78 & 0.62  &16.74&0.62 \\
URetinex-Net  &19.84&0.87 &-&-\\
UHDFour& 23.09 &0.87 &21.78&0.87\\
\shline
\textbf{PPTformer}&\textbf{25.48} & \textbf{0.93} &\textbf{23.52}& \textbf{0.91} \\
\shline
\end{tabular}}
% \vspace{-3mm}
\caption{Low-light image enhancement results on LOL~\cite{retinexnet_wei_bmvc18} and LOL-v2~\cite{lol-v2}.
} 
\label{table:Low-light image enhancement}
\end{center}
% \vspace{-5mm}
\end{table}

\subsection{Ablation study}\label{sec:Analysis and Discussion}
We conduct the ablation study using the low-light image enhancement model trained on LOL dateset~\cite{gopro2017} for $100,000$ iterations only. 
Except for reporting SSIM and LPIPS metrics, we also provide the FLOPs and the number of parameters  (Params) for reference, where FLOPs are computed on image size $256$$\times$$256$. 
Next, we describe the influence of each component individually.

\begin{figure}[!t]
% \vspace{-1mm}
\begin{center}
\begin{tabular}{cccccccccc}
\includegraphics[width = 0.2375\linewidth]{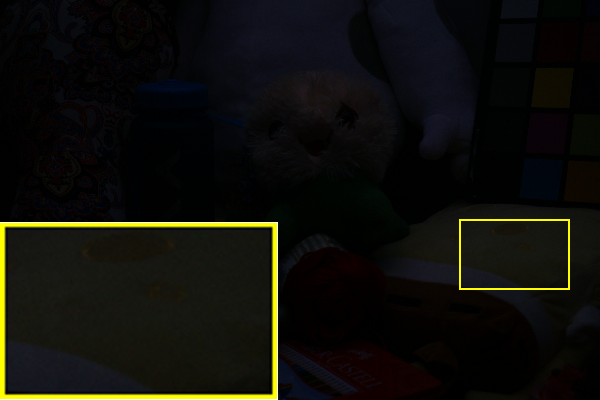}&\hspace{-4.5mm}
\includegraphics[width = 0.2375\linewidth]{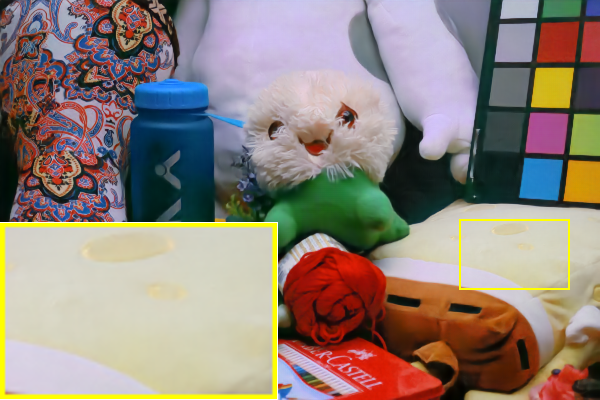}&\hspace{-4.5mm}
\includegraphics[width = 0.2375\linewidth]{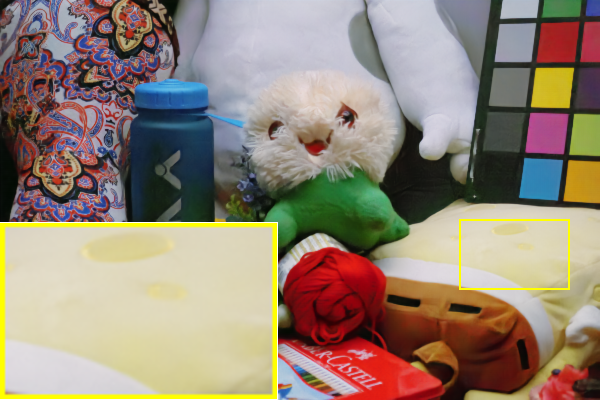}&\hspace{-4.5mm}
\includegraphics[width = 0.2375\linewidth]{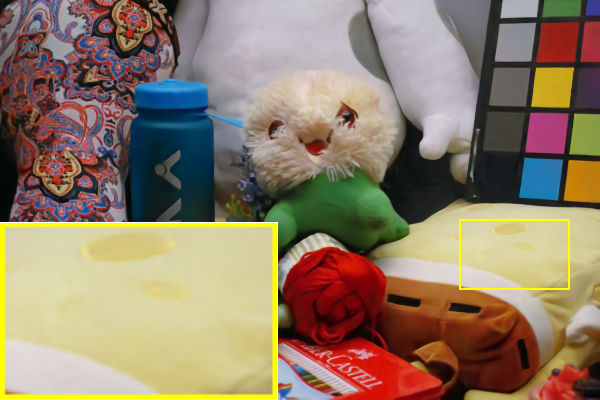}
\\
 (a) Input  &\hspace{-4.5mm} (b) Tab.~\ref{table: Effect on Mask}(a)  &\hspace{-4.5mm}   (c) Tab.~\ref{table: Effect on Mask}(b) &\hspace{-4.5mm}  (d) \textbf{Ours}
\\
\end{tabular}
% \vspace{-3mm}
\caption{Visual effect on parser.
Using the parser helps produce more natural results (d) with vivid colors than the model without using the parser (b) or with the degraded image (c).
}
\label{fig: Visual effect on mask.}
\end{center}
% \vspace{-3mm}
\end{figure}

{\flushleft \bf Effect of Parser.}
The main design of our PPTformer is that we use the parser map generated by a large visual foundation model~\cite{kirillov2023segany} to guide image restoration.
One may wonder to know that if this strategy is more effective than a previous widely-used scheme that uses input images as the restoration guidance or without using any conditions.
To answer this question, we conduct ablation experiments in Tab.~\ref{table: Effect on Mask}.
One can observe that when we do not integrate the parser to the restoration network, i.e., Tab.~\ref{table: Effect on Mask}(a), or we replace the parser with degraded images, i.e., Tab.~\ref{table: Effect on Mask}(b), the performance suffers from drop in terms of distortion metrics like SSIM and perceptual measurement like LPIPS, compared with our model with SAM parser guidance (Tab.~\ref{table: Effect on Mask}(c)).
Fig.~\ref{fig: Visual effect on mask.} presents a visual example, where our method is able to produce results with more consistent colors (Fig.~\ref{fig: Visual effect on mask.}(d)) than the model without using parser (Fig.~\ref{fig: Visual effect on mask.}(b)) or with degraded images as conditions (Fig.~\ref{fig: Visual effect on mask.}(c)).
%
% This also demonstrates that using SAM~\cite{kirillov2023segany} to parse degraded images to guide the restoration process facilitates restoration results with more consistent structures.
\begin{table}[!t]
\begin{center}
\setlength{\tabcolsep}{2pt}
\scalebox{0.8}{
\begin{tabular}{l|cc|cccccccccccccc}
\shline
 Experiment & SSIM~$\uparrow$ & LPIPS~$\downarrow$ & FLOPs (G) & Params (M)\\
\shline
\textbf{(a)} {w/o}~{Using Parser} & 0.9153 & 0.1211& 48.42 & 9.27 \\
\textbf{(b)} Parser$\rightarrow$Degraded Image & 0.9207 & 0.1151& 168.91 & 20.48 \\
% \cdashline{1-4}[3pt/2.5pt]
\shline
\textbf{(c)} \textbf{Ours} & \textbf{0.9248} &\textbf{0.1138}& 168.91 & 20.48 \\
\shline
\end{tabular}}
\end{center}
% \vspace{-3mm}
\caption{Effect on using of parser.
Compared with models without using the parser and with widely-used input degraded images as conditional modulation, our method that uses the parser to prompt the image restoration network performs the best. 
}
\label{table: Effect on Mask}
% \vspace{-3mm}
\end{table}

\begin{figure}[!t]
% \vspace{-4mm}
\begin{center}
\begin{tabular}{cccccccccc}
\includegraphics[width = 0.3175\linewidth, height= 0.2275\linewidth]{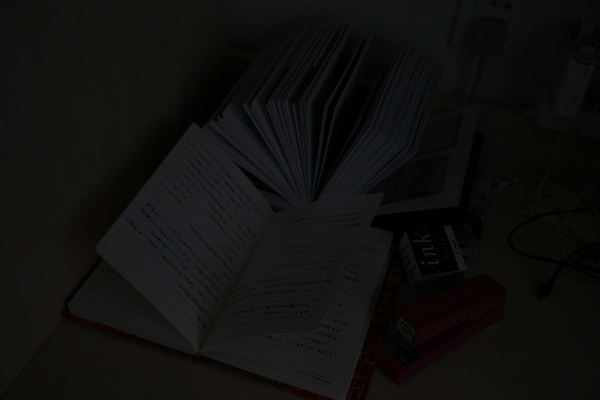}&\hspace{-4.5mm}
\includegraphics[width = 0.3175\linewidth, height= 0.2275\linewidth]{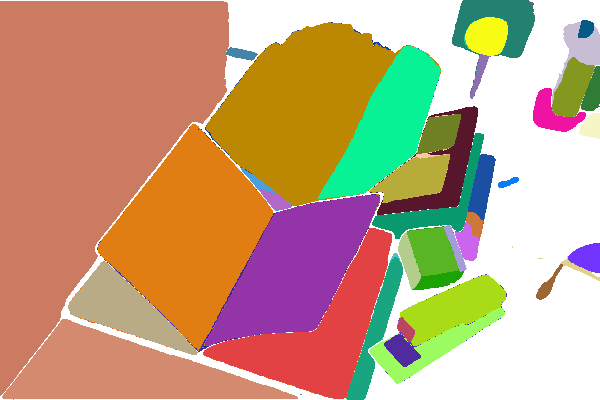}&\hspace{-4.5mm}
\includegraphics[width = 0.3175\linewidth, height= 0.2275\linewidth]{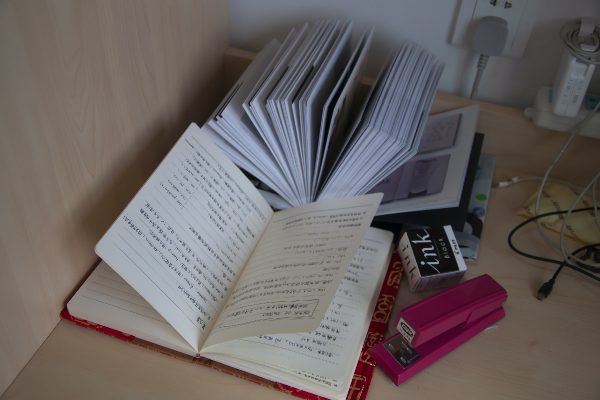}
\\
(a) Input  &\hspace{-4.5mm}  (b) Parser of (a)&\hspace{-4.5mm}    (c)  GT
\\
\includegraphics[width = 0.3175\linewidth, height= 0.2275\linewidth]{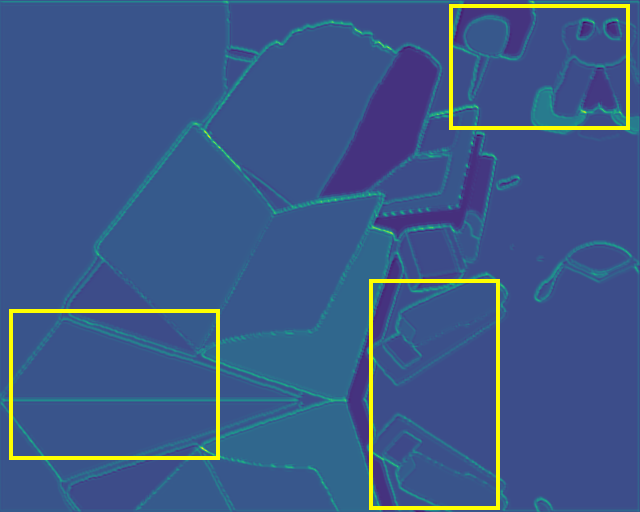}&\hspace{-4.5mm}
\includegraphics[width = 0.3175\linewidth, height= 0.2275\linewidth]{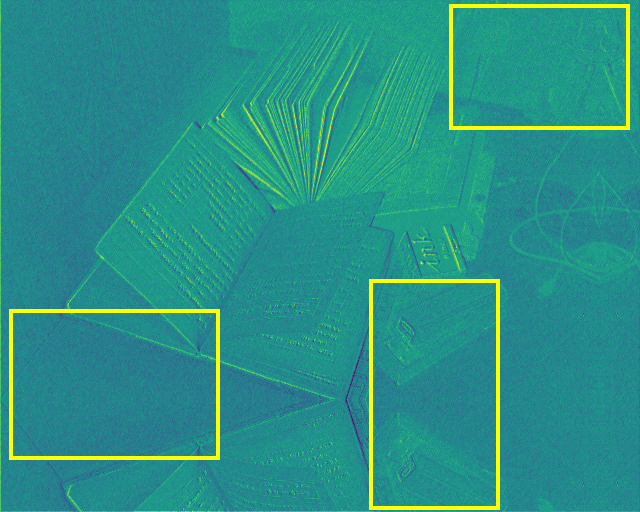}&\hspace{-4.5mm}
\includegraphics[width = 0.3175\linewidth, height= 0.2275\linewidth]{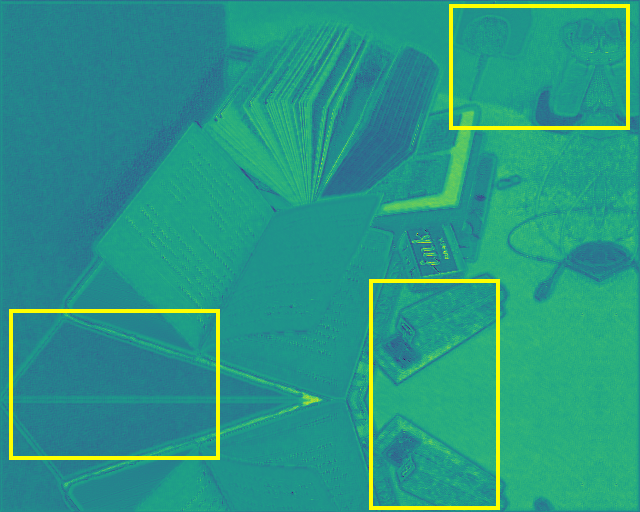}
\\
 (d) Parser Feature  &\hspace{-4.5mm} (e) Before   &\hspace{-4.5mm}    (f)  After 
\\
\end{tabular}
% \vspace{-3mm}
\caption{Visualization Understanding of the parser for guiding restoration in deep feature space.
We present the averaged channel-wise features at the first IN2PPT, before and after the application of IntraPPA and InterPPA to restoration features. This comparison notably illustrates how the parser, by offering valuable structural features (d), enhances the restoration features (e), leading to significantly sharper content (f) after being prompted through the use of IntraPPA and InterPPA.
}
\label{fig: Visualization Understanding of Parser for Guiding Restoration in Deep Feature Space.}
\end{center}
% \vspace{-3mm}
\end{figure}

To better comprehend the impact of our parser on restoration, we visualize the parser features and the restoration features both before and after applying Intra and Inter Parser-Prompted Attention, as shown in Fig.~\ref{fig: Visualization Understanding of Parser for Guiding Restoration in Deep Feature Space.}.
From Fig.~\ref{fig: Visualization Understanding of Parser for Guiding Restoration in Deep Feature Space.}(d), it is evident that the parser features prominently display salient structures. However, prior to the use of IntraPPA and InterPPA, the restoration features appear notably indistinct, as seen in Fig.~\ref{fig: Visualization Understanding of Parser for Guiding Restoration in Deep Feature Space.}(e). Post-application of the parser's prompting, the feature becomes significantly sharper (Fig.~\ref{fig: Visualization Understanding of Parser for Guiding Restoration in Deep Feature Space.}(f)), which greatly aids in restoration.
These visualizations offer profound insights into our PPTformer. They demonstrate that the parser generated by the SAM~\cite{kirillov2023segany} effectively provides valuable information, steering image restoration towards more refined results.

\begin{figure}[!t]
% \vspace{-4mm}
\begin{center}
\begin{tabular}{cccccccccc}
\includegraphics[width = 0.3175\linewidth]{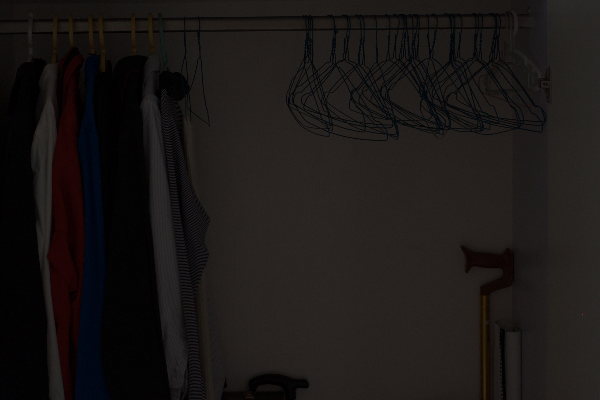}&\hspace{-4.5mm}
\includegraphics[width = 0.3175\linewidth]{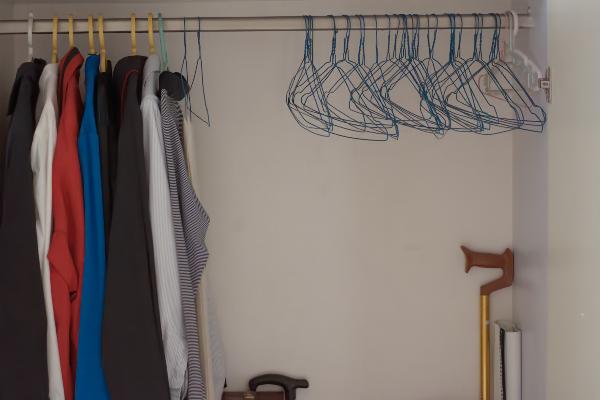}&\hspace{-4.5mm}
\includegraphics[width = 0.3175\linewidth]{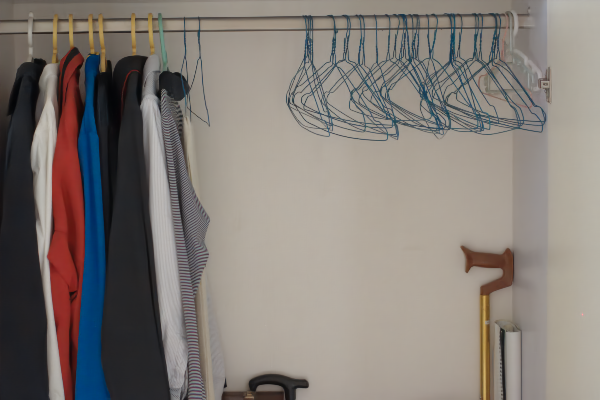}
\\
 (a) Input &\hspace{-4.5mm}  (b) Tab.~\ref{table: Effect on Inter and Intra Mask-Prompted Attention}(a) &\hspace{-4.5mm}   (c)  \textbf{Ours}
\\
\end{tabular}
% \vspace{-3mm}
\caption{Visual effect on intra and inter parser-prompted attention.
Using IntraPPA and InterPPA helps produce more natural results with vivid colors, i.e., (c), than the model without both IntraPPA and InterPPA, i.e., (b).
}
\label{fig: Visual effect on inter and intra mask-prompted attention.}
\end{center}
% \vspace{-5mm}
\end{figure}

{\flushleft \bf Effect of Intra and Inter Parser-Prompted Attention.}
We propose IntraPPA and InterPPA to implicitly and explicitly explore the parser to guide image restoration. 
Hence, analyzing this module is necessary.
Tab.~\ref{table: Effect on Inter and Intra Mask-Prompted Attention} summarises the ablation results.
As can be seen, removing IntraPPA or InterPPA or both of them decreases the enhancement performance.
On the other hand, we also use two IntraPPA or two InterPPA to replace the IntraPPA-InterPPA structure.
However, we note that our default model achieves the best, adequately demonstrating the effectiveness of our proposed IntraPPA and InterPPA.
The visual example presented in Fig.~\ref{fig: Visual effect on inter and intra mask-prompted attention.} further suggests that using IntraPPA-InterPPA helps produce more natural results.

\begin{table}[!t]
% \vspace{-5mm}
\begin{center}
\setlength{\tabcolsep}{2pt}
\scalebox{0.81}{
\begin{tabular}{l|cc|cccccccccccccc}
\shline
 Experiment & SSIM~$\uparrow$ & LPIPS~$\downarrow$ & FLOPs (G) & Params (M)\\
\shline
\textbf{(a)} w/o IntraPPA\&InterPPA& 0.9110 & 0.1383& 129.05 &17.37 \\
\textbf{(b)} w/o IntraPPA &  0.9152 &0.1322&158.75&19.69 \\
\textbf{(c)} w/o InterPPA &0.9157 &0.1305 &164.29  & 20.12  \\
\textbf{(d)} Both IntraPPA &0.9176 & 0.1200& 163.38 &20.05 \\
\textbf{(e)} Both InterPPA & 0.9164 &0.1230& 174.45 &20.90 \\
% \hdashline
\shline
\textbf{(f)} \textbf{Ours} & \textbf{0.9248} & \textbf{0.1138}& 168.91 & 20.48 \\
\shline
\end{tabular}}
\end{center}
% \vspace{-3mm}
\caption{Effect on intra and inter parser-prompted attention.
Compared with models without intra and inter attention or with one of them or with both intra and inter attention, our method that utilizes intra and inter attention can implicitly and explicitly learn useful content from the parser, thus leading to the best results.
}
\label{table: Effect on Inter and Intra Mask-Prompted Attention}
% \vspace{-5mm}
\end{table}

\begin{figure}[!t]
\begin{center}
\begin{tabular}{cccccccccc}
\includegraphics[width = 0.3175\linewidth]{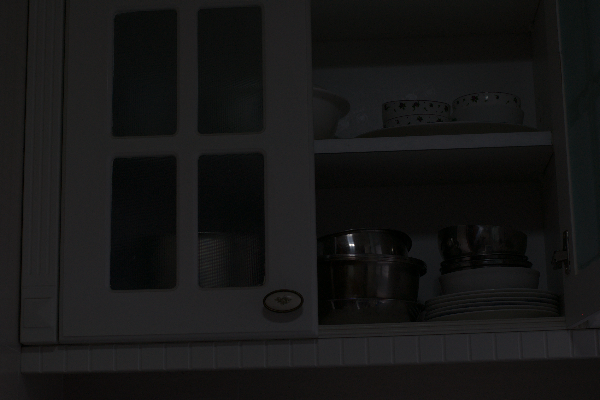}&\hspace{-4.5mm}
\includegraphics[width = 0.3175\linewidth]{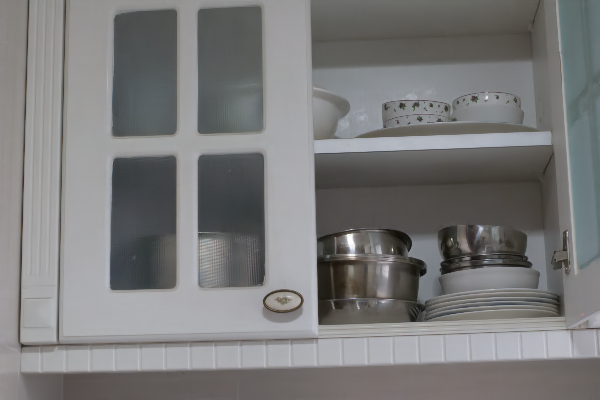}&\hspace{-4.5mm}
\includegraphics[width = 0.3175\linewidth]{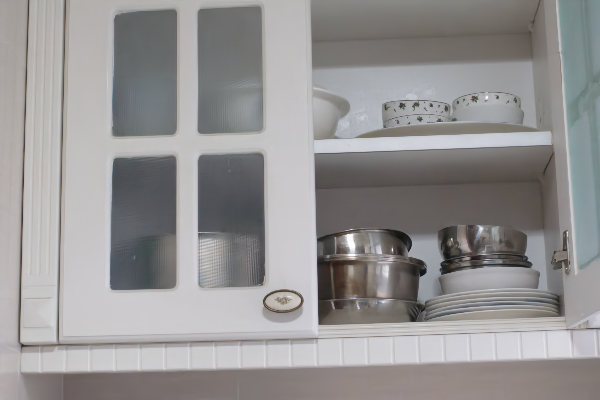}
\\
 (a) Input &\hspace{-4.5mm}  (b) SFT&\hspace{-4.5mm}    (c)  \textbf{Ours}
\\
\end{tabular}
% \vspace{-3mm}
\caption{Visual effect on bidirectional parser-prompted fusion.
Our BiPPF helps recover clearer results with more natural colors than the model with widely-used fusion module SFT~\cite{Wang_2018_CVPR}.
}
\label{fig: Visual effect on bidirectional mask-prompted fusion.}
\end{center}
% \vspace{-3mm}
\end{figure}

{\flushleft \bf Effect of Bidirectional Parser-Prompted Fusion.}
As we introduce the BiPPF module to fuse the parser features and restoration ones, we need to compare it with the widely-used feature fuse module SFT~\cite{Wang_2018_CVPR}.
Tab.~\ref{table: Effect on Bidirectional Mask-Prompted Fusion} shows that our proposed BiPPF significantly outperforms the SFT in terms of SSIM and LPIPS. 
Fig.~\ref{fig: Visual effect on bidirectional mask-prompted fusion.} shows our BiPPF helps produce more vivid results than the model with SFT~\cite{Wang_2018_CVPR}.

%%%%%%%%%%%%%%%%%%%%%%%%%%%
% \subsection{Discussion and Limitations}
{\flushleft \bf Discussion and Limitations.}
To efficiently train the PPTformer, we initially generate the parser using SAM~\cite{kirillov2023segany} offline. This involves first leveraging SAM to produce the parser, which is then input into the networks as an image. We acknowledge that using SAM's intermediate features might yield better restoration results than our current method. However, incorporating SAM directly into the training process would greatly increase computational demands, adversely affecting training efficiency. Additionally, while our ablation experiments confirm SAM's advantages in low-light image enhancement over models that either do not use the parser or use degraded images as input conditions, its effectiveness in other image restoration tasks like dynamic scene deblurring and image dehazing remains to be further investigated.

\begin{table}[!t]
% \vspace{-6mm}
\begin{center}
\setlength{\tabcolsep}{3pt}
\scalebox{0.91}{
\begin{tabular}{l|cc|cccccccccccccc}
\shline
 Experiment & SSIM~$\uparrow$ & LPIPS~$\downarrow$ & FLOPs (G) & Params (M)\\
\shline
\textbf{(a)} BiPPF$\rightarrow$SFT  &0.9154  & 0.1305&  120.59&14.39 \\
\shline
\textbf{(b)} \textbf{Ours}  & \textbf{0.9248} & \textbf{0.1138}& 168.91 & 20.48 \\
\shline
\end{tabular}}
\end{center}
% \vspace{-3mm}
\caption{Effect on bidirectional parser-prompted fusion.
Compared with the widely-used feature fusion module SFT~\cite{Wang_2018_CVPR}, our proposed BiPPF significantly outperforms it in terms of SSIM and LPIPS, demonstrating the effectiveness of BiPPF.
}
\label{table: Effect on Bidirectional Mask-Prompted Fusion}
% \vspace{-3mm}
\end{table}

%%%%%%%%%%%%%%%%%%%%%%%%%%%

\section{Concluding Remarks}
We have proposed the PPTformer, an inter and intra Parser-Prompted Transformer, for image restoration.
Our PPTformer nicely integrates the power of visual foundation models into the restoration process.
To effectively fuse the parser generated by SAM~\cite{kirillov2023segany} into restoration, we have proposed the inter and intra parser-prompted attention to implicitly and explicitly learn useful information to facilitate restoration.
Moreover, we have suggested a bidirectional parser-prompted fusion scheme to better fuse parser features with restoration ones.
Extensive experiments have demonstrated that our PPTformer outperforms state-of-the-art approaches on 4 restoration tasks, including image deraining, single-image defocus deblurring, image desnowing, and low-light image enhancement.

\section*{Acknowledgements}
% We thank the anonymous reviewers for their constructive suggestions.
This work was supported by the National Natural Science Foundation of China (No.62306343), the China Postdoctoral Science Foundation (No.2024M753741), the Centre for Advances in Reliability and Safety (CAiRS) admitted under AiR@InnoHK Research Cluster. 

\bibliography{aaai25}
% \newpage
% \clearpage
% \input{Checklist}
\end{document}